\documentclass[acmsmall]{acmart}
\usepackage{microtype}
\usepackage{multirow}
\usepackage{makecell}
\usepackage{booktabs}
\usepackage{arydshln}
\usepackage{hhline}
\usepackage{enumitem}
\RequirePackage{color,graphicx}
\usepackage{xcolor}
\usepackage{hyperref}
\definecolor{linkcolour}{rgb}{0,0.2,0.6}
\hypersetup{colorlinks=true,breaklinks,urlcolor=linkcolour, citecolor=linkcolour,linkcolor=linkcolour,allcolors=linkcolour}
\usepackage[noabbrev,nameinlink]{cleveref}

\let\oldhref\href
\renewcommand{\href}[3][ACMPurple]{\oldhref{#2}{\color{#1}{#3}}}
\renewcommand{\cref}[2][ACMPurple]{\textcolor{#1}{\ref{#2}}}

\AtBeginDocument{%
  \providecommand\BibTeX{{%
    \normalfont B\kern-0.5em{\scshape i\kern-0.25em b}\kern-0.8em\TeX}}}

\setcopyright{rightsretained}
\copyrightyear{2022}
\acmYear{2022}
\acmDOI{XXXXXXX.XXXXXXX}

\acmJournal{CSUR}
\acmVolume{1}
\acmNumber{1}
\acmArticle{1}
\acmMonth{10}

\begin{document}

\title{Foundations \& Trends in Multimodal Machine Learning: Principles, Challenges, and Open Questions}

\renewcommand{\shorttitle}{Foundations \& Trends in Multimodal Machine Learning}

\author{Paul Pu Liang}
\email{pliang@cs.cmu.edu}
\author{Amir Zadeh}
\email{abagherz@cs.cmu.edu}
\author{Louis-Philippe Morency}
\email{morency@cs.cmu.edu}
\affiliation{%
  \institution{\\Machine Learning Department and Language Technologies Institute, Carnegie Mellon University}
  \streetaddress{5000 Forbes Ave}
  \city{Pittsburgh}
  \state{PA}
  \country{USA}
  \postcode{15213}
}


\begin{abstract}
Multimodal machine learning is a vibrant multi-disciplinary research field that aims to design computer agents with intelligent capabilities such as understanding, reasoning, and learning through integrating multiple communicative modalities, including linguistic, acoustic, visual, tactile, and physiological messages. With the recent interest in video understanding, embodied autonomous agents, text-to-image generation, and multisensor fusion in application domains such as healthcare and robotics, multimodal machine learning has brought unique computational and theoretical challenges to the machine learning community given the heterogeneity of data sources and the interconnections often found between modalities. However, the breadth of progress in multimodal research has made it difficult to identify the common themes and open questions in the field. By synthesizing a broad range of application domains and theoretical frameworks from both historical and recent perspectives, this paper is designed to provide an overview of the computational and theoretical foundations of multimodal machine learning. We start by defining three key principles of modality \textit{heterogeneity}, \textit{connections}, and \textit{interactions} that have driven subsequent innovations, and propose a taxonomy of six core technical challenges: \textit{representation}, \textit{alignment}, \textit{reasoning}, \textit{generation}, \textit{transference}, and \textit{quantification} covering historical and recent trends. Recent technical achievements will be presented through the lens of this taxonomy, allowing researchers to understand the similarities and differences across new approaches. We end by motivating several open problems for future research as identified by our taxonomy.
\end{abstract}

\ccsdesc[500]{Computing methodologies~Machine learning}
\ccsdesc[500]{Computing methodologies~Artificial intelligence}
\ccsdesc[300]{Computing methodologies~Computer vision}
\ccsdesc[300]{Computing methodologies~Natural language processing}

\keywords{multimodal machine learning, representation learning, data heterogeneity, feature interactions, language and vision, multimedia}


\maketitle

\vspace{-2mm}
\section{Introduction}
\vspace{-1mm}
It has always been a grand goal of artificial intelligence to develop computer agents with intelligent capabilities such as understanding, reasoning, and learning through multimodal experiences and data, similar to how humans perceive and interact with our world using multiple sensory modalities. With recent advances in embodied autonomous agents~\citep{brodeur2017home,savva2019habitat}, self-driving cars~\citep{xiao2020multimodal}, image and video understanding~\citep{alayrac2022flamingo,sun2019videobert}, image and video generation~\citep{ramesh2021zero,singer2022make}, and multisensor fusion in application domain such as robotics~\citep{lee2019making,marge2022spoken} and healthcare~\citep{MIMIC,liang2021multibench}, we are now closer than ever to intelligent agents that can integrate and learn from many sensory modalities.
This vibrant multi-disciplinary research field of multimodal machine learning brings unique challenges given the heterogeneity of the data and the interconnections often found between modalities, and has widespread applications in multimedia~\citep{1667983}, affective computing~\citep{PORIA201798}, robotics~\citep{kirchner2019embedded,lee2019making}, human-computer interaction~\citep{obrenovic2004modeling,sharma2002toward}, and healthcare~\citep{cai2019survey,muhammad2021comprehensive}.

However, the rate of progress in multimodal research has made it difficult to identify the common themes underlying historical and recent work, as well as the key open questions in the field. By synthesizing a broad range of multimodal research, this paper is designed to provide an overview of the methodological, computational, and theoretical foundations of multimodal machine learning, which complements recent application-oriented surveys in vision and language~\citep{uppal2022multimodal}, language and reinforcement learning~\citep{luketina2019survey}, multimedia analysis~\citep{atrey2010multimodal}, and human-computer interaction~\citep{jaimes2007multimodal}.

\begin{figure}[t]
\centering
\vspace{-0mm}
\includegraphics[width=0.8\linewidth]{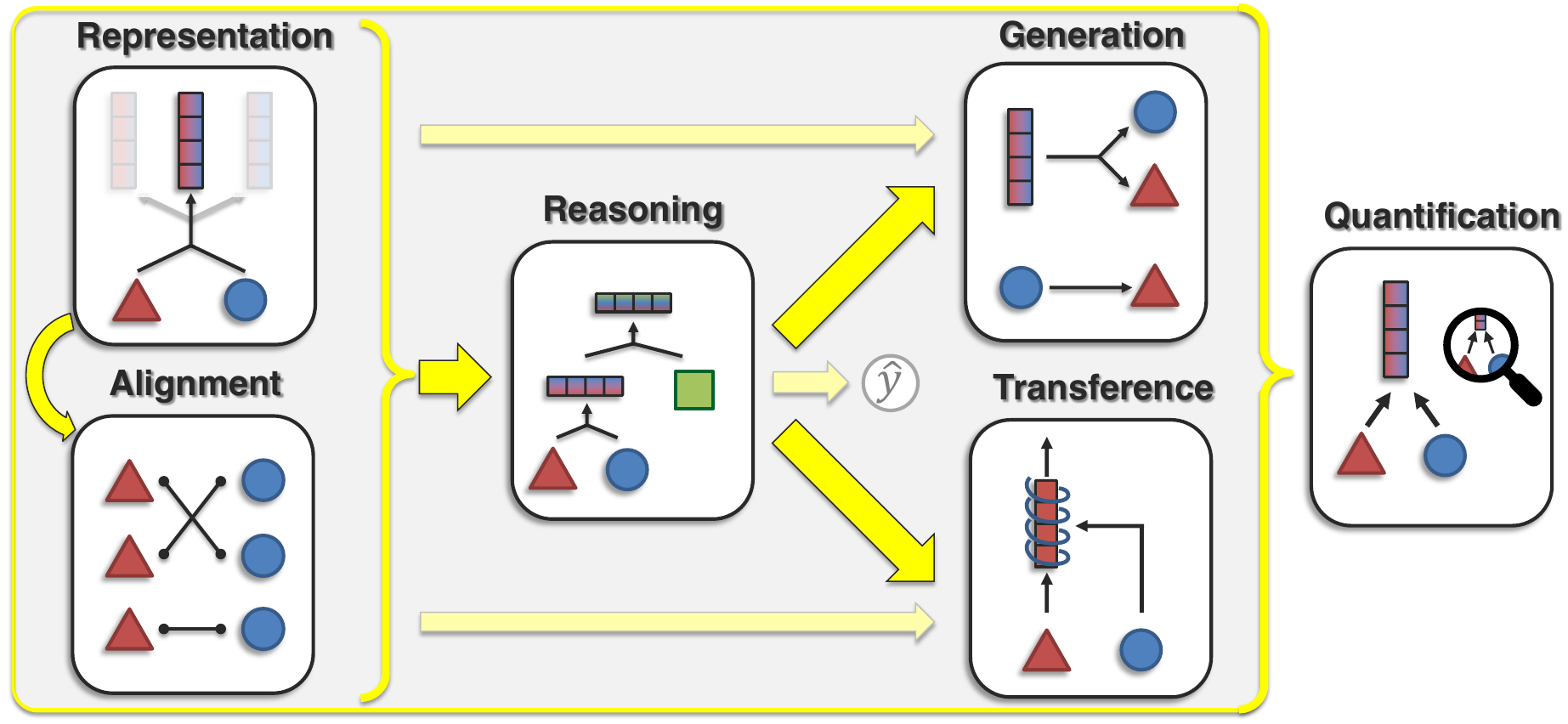}
\vspace{-2mm}
\caption{Core research challenges in multimodal learning: (1) \textit{Representation} studies how to represent and summarize multimodal data to reflect the heterogeneity and interconnections between individual modality elements. (2) \textit{Alignment} aims to identify the connections and interactions across all elements. (3) \textit{Reasoning} aims to compose knowledge from multimodal evidence usually through multiple inferential steps for a task. (4) \textit{Generation} involves learning a generative process to produce raw modalities that reflect
cross-modal interactions, structure, and coherence. (5) \textit{Transference} aims to transfer knowledge between modalities and their representations. (6) \textit{Quantification} involves empirical and theoretical studies to better understand the multimodal learning process.}
\label{fig:intro_challenges}
\vspace{-4mm}
\end{figure}

To better understand the foundations of multimodal machine learning, we begin by defining (in \S\cref{sec:definitions}) three key principles that have driven subsequent technical challenges and innovations: (1) modalities are \textit{heterogeneous} because the information present often shows diverse qualities, structures, and representations, (2) modalities are \textit{connected} since they are often related and share commonalities, and (3) modalities \textit{interact} to give rise to new information when used for task inference.
Building upon these definitions, we propose a new taxonomy of six core challenges in multimodal learning: \textit{representation}, \textit{alignment}, \textit{reasoning}, \textit{generation}, \textit{transference}, and \textit{quantification} (see Figure~\cref{fig:intro_challenges}). These constitute core multimodal technical challenges that are understudied in conventional unimodal machine learning, and need to be tackled in order to progress the field forward:
\begin{enumerate}[noitemsep,topsep=0pt,nosep,leftmargin=*,parsep=0pt,partopsep=0pt]
    \item \textbf{Representation (\S\cref{sec:representation}):} Can we learn representations that reflect heterogeneity and interconnections between modality elements? We will cover approaches for (1) \textit{representation fusion}: integrating information from two or more modalities to capture cross-modal interactions, (2) \textit{representation coordination}: interchanging cross-modal information with the goal of keeping the same number of representations but improving multimodal contextualization, and (3) \textit{representation fission}: creating a larger set of disjoint representations that reflects knowledge about internal structure such as data clustering or factorization.

    \item \textbf{Alignment (\S\cref{sec:alignment}):} How can we identify the connections and interactions between modality elements? Alignment is challenging since it may depend on long-range dependencies, involves ambiguous segmentation (e.g., words or utterances), and could be either one-to-one, many-to-many, or not exist at all. We cover (1) \textit{discrete alignment}: identifying connections between discrete elements across modalities, (2) \textit{continuous alignment}: modeling alignment between continuous modality signals with ambiguous segmentation, and (3) \textit{contextualized representations}: learning better representations by capturing cross-modal interactions between elements.

    \item \textbf{Reasoning (\S\cref{sec:reasoning})} is defined as composing knowledge, usually through multiple inferential steps, that exploits the problem structure for a specific task. Reasoning involves (1) \textit{modeling the structure} over which composition occurs, (2) the \textit{intermediate concepts} in the composition process, (3) understanding the \textit{inference paradigm} of more abstract concepts, and (4) leveraging large-scale \textit{external knowledge} in the study of structure, concepts, and inference.

    \item \textbf{Generation (\S\cref{sec:generation})} involves learning a generative process to produce raw modalities. We categorize its subchallenges into (1) \textit{summarization}: summarizing multimodal data to reduce information content while highlighting the most salient parts of the input, (2) \textit{translation}: translating from one modality to another and keeping information content while being consistent with cross-modal connections, and (3) \textit{creation}: simultaneously generating multiple modalities to increase information content while maintaining coherence within and across modalities.
    
    \item \textbf{Transference (\S\cref{sec:transference})} aims to transfer knowledge between modalities, usually to help the target modality, which may be noisy or with limited resources. Transference is exemplified by (1) \textit{cross-modal transfer}: adapting models to tasks involving the primary modality, (2) \textit{co-learning}: transferring information from secondary to primary modalities by sharing representation spaces between both modalities, and (3) \textit{model induction}: keeping individual unimodal models separate but transferring information across these models.
    
    \item \textbf{Quantification (\S\cref{sec:quantification}):} The sixth and final challenge involves empirical and theoretical studies to better understand (1) the dimensions of \textit{heterogeneity} in multimodal datasets and how they subsequently influence modeling and learning, (2) the presence and type of modality \textit{connections and interactions} in multimodal datasets and captured by trained models, and (3) the \textit{learning} and optimization challenges involved with heterogeneous data.
\end{enumerate}

\vspace{1mm}
Finally, we conclude this paper with a long-term perspective in multimodal learning by motivating open research questions identified by this taxonomy. This survey was also presented by the authors in a visual medium through tutorials at \href{https://cmu-multicomp-lab.github.io/mmml-tutorial/cvpr2022/}{CVPR 2022} and \href{https://cmu-multicomp-lab.github.io/mmml-tutorial/naacl2022/}{NAACL 2022}, as well as courses \href{https://cmu-multicomp-lab.github.io/mmml-course/fall2022/}{11-777 Multimodal Machine Learning} and \href{https://cmu-multicomp-lab.github.io/adv-mmml-course/spring2023/}{11-877 Advanced Topics in Multimodal Machine Learning} at CMU. The reader is encouraged to check out these publicly available video recordings, additional reading materials, and discussion probes motivating open research questions in multimodal learning.

\vspace{-2mm}
\section{Foundational Principles in Multimodal Research}
\label{sec:definitions}
\vspace{-0mm}

A \textit{modality} refers to a way in which a natural phenomenon is perceived or expressed. For example, modalities include speech and audio recorded through microphones, images and videos captured via cameras, and force and vibrations captured via haptic sensors.
Modalities can be placed along a spectrum from \textit{raw} to \textit{abstract}: raw modalities are those more closely detected from a sensor, such as speech recordings from a microphone or images captured by a camera. Abstract modalities are those farther away from sensors, such as language extracted from speech recordings, objects detected from images, or even abstract concepts like sentiment intensity and object categories.

\textit{Multimodal} refers to situations where multiple modalities are involved. From a research perspective, multimodal entails the computational study of \textit{heterogeneous} and \textit{interconnected} (connections + interactions) modalities. Firstly, modalities are \textit{heterogeneous} because the information present in different modalities will often show diverse qualities, structures, and representations. Secondly, these modalities are not independent entities but rather share \textit{connections} due to complementary information. Thirdly, modalities \textit{interact} in different ways when they are integrated for a task. We expand on these three foundational principles of multimodal research in the following subsections.

\begin{figure}[t]
\centering
\vspace{-1mm}
\includegraphics[width=0.8\linewidth]{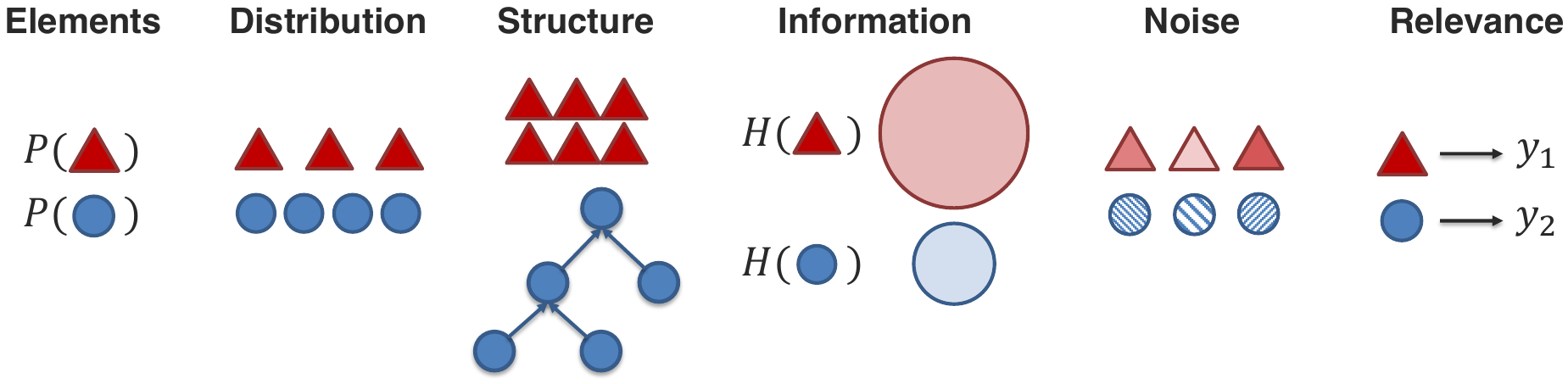}
\vspace{-3mm}
\caption{The information present in different modalities will often show diverse qualities, structures, and representations. \textbf{Dimensions of heterogeneity} can be measured via differences in individual elements and their distribution, the structure of elements, as well as modality information, noise, and task relevance.}
\label{fig:hetero}
\vspace{-4mm}
\end{figure}

\vspace{-2mm}
\subsection{Principle 1: Modalities are Heterogeneous}
\vspace{-1mm}

The principle of heterogeneity reflects the observation that the information present in different modalities will often show diverse qualities, structures, and representations. Heterogeneity should be seen as a spectrum: two images from the same camera which capture the same view modulo camera wear and tear are closer to homogeneous, two different languages which capture the same meaning but are different depending on language families are slightly heterogeneous, language and vision are even more heterogeneous, and so on.
In this section, we present a non-exhaustive list of dimensions of heterogeneity (see Figure~\cref{fig:hetero} for an illustration). These dimensions are complementary and may overlap; each multimodal problem likely involves heterogeneity in multiple dimensions.

\begin{enumerate}[noitemsep,topsep=0pt,nosep,leftmargin=*,parsep=0pt,partopsep=0pt]
    \item \textbf{Element representation}: Each modality is typically comprised of a set of elements - the most basic unit of data which cannot (or rather, the user chooses to not) be broken down into further units~\citep{barthes1977image,liang2022brainish}. For example, typed text is recorded via a set of characters, videos are recorded via a set of frames, and graphs are recorded via a set of nodes and edges. What are the basic elements present in each modality, and how can we represent them? Formally, this dimensions measures heterogeneity in the sample space or representation space of modality elements.

    \item \textbf{Distribution} refers to the frequency and likelihood of elements in modalities. Elements typically follow a unique distribution, with words in a linguistic corpus following Zipf's Law as a classic example. Distribution heterogeneity then refers to the differences in frequencies and likelihoods of elements, such as different frequencies in recorded signals and the density of elements.

    \item \textbf{Structure}: Natural data exhibits structure in the way individual elements are composed to form entire modalities~\citep{bronstein2021geometric}. For example, images exhibit spatial structure across individual object elements, language is hierarchically composed of individual words, and signals exhibit temporal structure across time. Structure heterogeneity refers to differences in this underlying structure.
    
    \item \textbf{Information} measures the total information content present in each modality. Subsequently, information heterogeneity measures the differences in information content across modalities, which could be formally measured by information theoretic metrics~\citep{shannon1948mathematical}.

    \item \textbf{Noise}: Noise can be introduced at several levels across naturally occurring data and also during the data recording process. Natural data noise includes occlusions, imperfections in human-generated data (e.g., imperfect keyboard typing or unclear speech), or data ambiguity due to sensor failures~\citep{liang2021multibench}. Noise heterogeneity measures differences in noise distributions across modalities, as well as differences in signal-to-noise ratio.

    \item \textbf{Relevance}: Finally, each modality shows different relevance toward specific tasks and contexts - certain modalities may be more useful for certain tasks than others~\citep{gat2021perceptual}. Task relevance describes how modalities can be used for inference, while context relevance describes how modalities are contextualized with other modalities.
\end{enumerate}
It is useful to take these dimensions of heterogeneity into account when studying both unimodal and multimodal data. In the unimodal case, specialized encoders are typically designed to capture these unique characteristics in each modality~\citep{bronstein2021geometric}. In the multimodal case, modeling heterogeneity is useful when learning representations and capturing alignment~\citep{zamir2018taskonomy}, and is a key subchallenge in quantifying multimodal models~\citep{liang2022highmmt}.

\vspace{-2mm}
\subsection{Principle 2: Modalities are Connected}
\vspace{-1mm}

\begin{figure}[t]
\centering
\vspace{-1mm}
\includegraphics[width=0.65\linewidth]{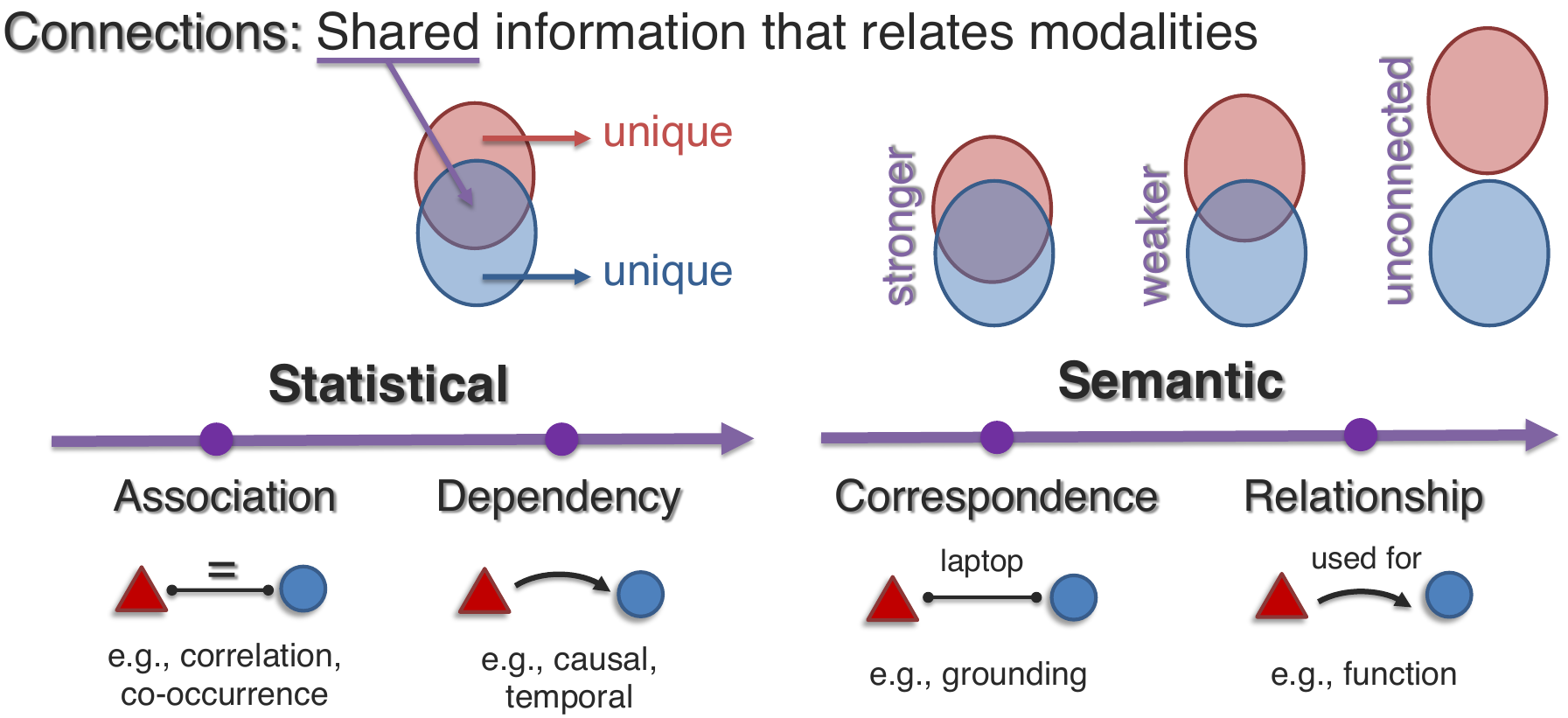}
\vspace{-3mm}
\caption{\textbf{Modality connections} describe how modalities are related and share commonalities, such as correspondences between the same concept in language and images or dependencies across spatial and temporal dimensions. Connections can be studied through both statistical and semantic perspectives.}
\label{fig:connections}
\vspace{-4mm}
\end{figure}

Although modalities are heterogeneous, they are often connected due to shared complementary information. The presence of \textit{shared} information is often in contrast to \textit{unique} information that exists solely in a single modality~\citep{williams2010nonnegative}. Modality connections describe the extent and dimensions in which information can be shared across modalities. When reasoning about the connections in multimodal data, it is helpful to think about both bottom-up (statistical) and top-down (semantic) approaches (see Figure~\cref{fig:connections}). From a statistical data-driven perspective, connections are identified from distributional patterns in multimodal data, while semantic approaches define connections based on our domain knowledge about how modalities share and contain unique information.
\begin{enumerate}[noitemsep,topsep=0pt,nosep,leftmargin=*,parsep=0pt,partopsep=0pt]
    \item \textbf{Statistical association} exists when the values of one variable relate to the values of another. For example, two elements may co-occur with each other, resulting in a higher frequency of both occurring at the same time. Statistically, this could lead to correlation - the degree in which elements are linearly related, or other non-linear associations. From a data-driven perspective, discovering which elements are associated with each other is important for modeling the joint distributions across modalities during multimodal representation and alignment~\citep{tian2020makes}.

    \item \textbf{Statistical dependence} goes deeper than association and requires an understanding of the exact type of statistical dependency between two elements. For example, is there a causal dependency from one element to another, or an underlying confounder causing both elements to be present at the same time? Other forms of dependencies could be spatial or temporal: one element occurring above the other, or after the other. Typically, while statistical association can be estimated purely from data, understanding the nature of statistical dependence requires some knowledge of the elements and their underlying relationships~\cite{nickel2015review,turney2005corpus}.
    
    \item \textbf{Semantic correspondence} can be seen as the problem of ascertaining which elements in one modality share the same semantic meaning as elements in another modality~\citep{otto2020characterization}. Identifying correspondences is fundamental in many problems related to language grounding~\cite{chandu2021grounding}, translation and retrieval~\citep{plummer2015flickr30k}, and cross-modal alignment~\citep{tan2019lxmert}.
    
    \item \textbf{Semantic relations}: Finally, semantic relations generalize semantic correspondences: instead of modality elements sharing the same exact meaning, semantic relations includes an attribute describing the exact nature of the relationship between two modality elements, such as semantic, logical, causal, or functional relations. Identifying these semantically related connections is important for higher-order reasoning~\citep{marsh2003taxonomy,barthes1977image}.
\end{enumerate}

\begin{figure}[t]
\centering
\vspace{-0mm}
\includegraphics[width=0.7\linewidth]{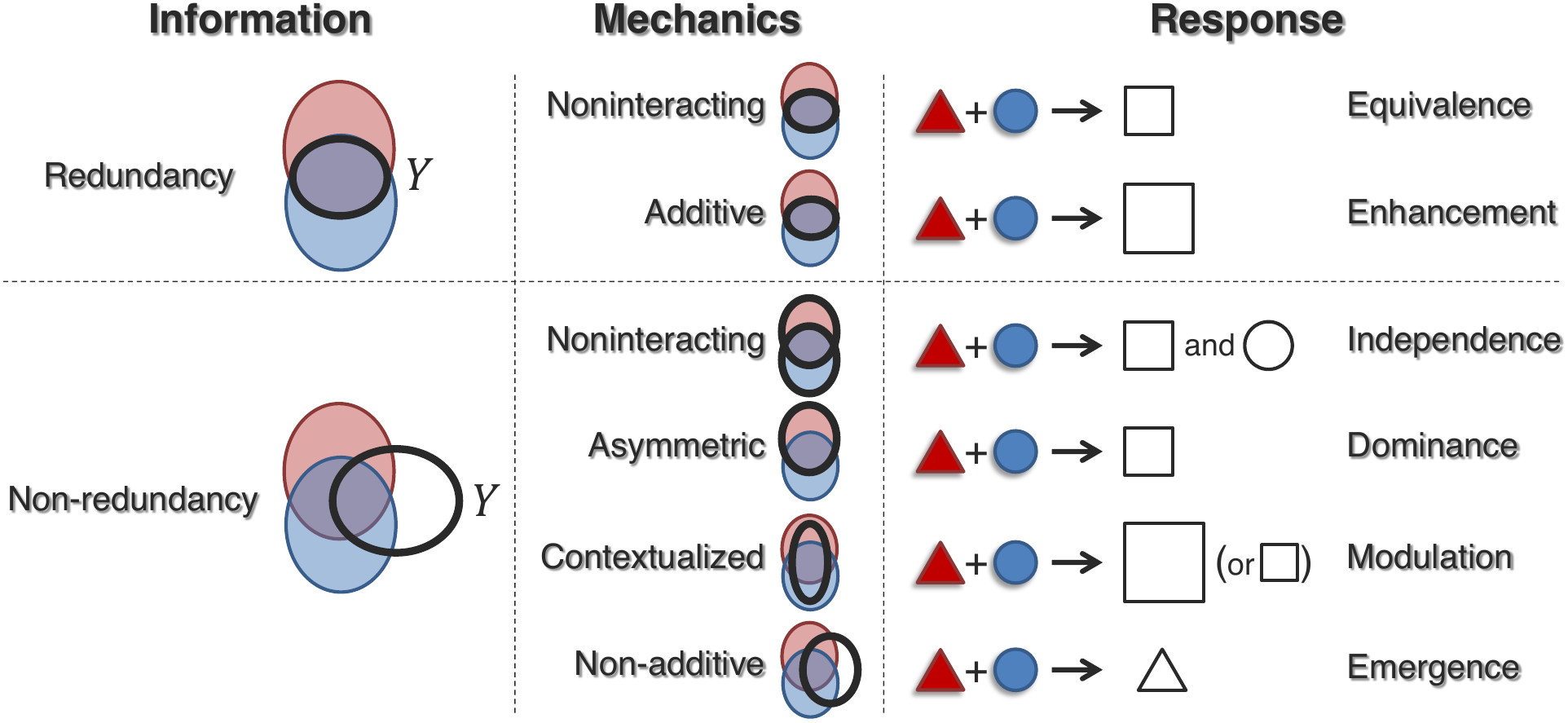}
\vspace{-2mm}
\caption{\textbf{Several dimensions of modality interactions}: (1) Interaction information studies whether common redundant information or unique non-redundant information is involved in interactions; (2) interaction mechanics study the manner in which interaction occurs, and (3) interaction response studies how the inferred task changes in the presence of multiple modalities.}
\label{fig:interactions}
\vspace{-4mm}
\end{figure}

\vspace{-2mm}
\subsection{Principle 3: Modalities Interact}
\vspace{-1mm}

Modality interactions study how modality elements interact to give rise to new information when integrated together for task \textit{inference}.
We note an important difference between modality connections and interactions: connections exist within multimodal data itself, whereas interactions only arise when modalities are integrated and processed together to bring a new response. In Figure~\cref{fig:interactions}, we provide a high-level illustration of some dimensions of interactions that can exist.
\begin{enumerate}[noitemsep,topsep=0pt,nosep,leftmargin=*,parsep=0pt,partopsep=0pt]
    \item \textbf{Interaction information} investigates the type of connected information that is involved in an interaction. When an interaction involves shared information common to both modalities, the interaction is \textit{redundant}, while a \textit{non-redundant} interaction is one that does not solely rely on shared information, and instead relies on different ratios of shared, unique, or possibly even synergistic information~\citep{williams2010nonnegative}.

    \item \textbf{Interaction mechanics} are the functional operators involved when integrating modality elements for task inference. For example, interactions can be expressed as statistically additive, non-additive, and non-linear forms~\citep{jayakumar2020multiplicative}, as well as from a semantic perspective where two elements interact through a logical, causal, or temporal operation~\citep{unsworth2014multimodality}.

    \item \textbf{Interaction response} studies how the inferred response changes in the presence of multiple modalities. For example, through sub-dividing redundant interactions, we can say that two modalities create an equivalence response if the multimodal response is the same as responses from either modality, or enhancement if the multimodal response displays higher confidence. On the other hand, non-redundant interactions such as modulation or emergence happen when there exist different multimodal versus unimodal responses~\cite{partan1999communication}.
\end{enumerate}

\vspace{-2mm}
\subsection{Core Technical Challenges}
\vspace{-1mm}

\begin{table*}[t]
\fontsize{9}{11}\selectfont
\setlength\tabcolsep{2.0pt}
\vspace{-0mm}
\caption{This table summarizes our taxonomy of $6$ core challenges in multimodal machine learning, their subchallenges, categories of corresponding approaches, and representative examples. We believe that this taxonomy can help to catalog rapid progress in this field and better identify the open research questions.}
\centering
\footnotesize
\vspace{0mm}
\begin{tabular}{c|c|ccccccc}
\hline \hline
Challenge & Subchallenge & Approaches \& key examples \\
\hline

\multirow{3}{*}{Representation (\S\cref{sec:representation})} & \multirow{1}{*}{Fusion (\S\cref{sec:representation1})} & Abstract~\cite{jayakumar2020multiplicative,zadeh2017tensor} \& raw~\cite{barnum2020benefits,rajagopalan2016extending} fusion \\
\Xcline{2-5}{0.5\arrayrulewidth}
& \multirow{1}{*}{Coordination (\S\cref{sec:representation2})} & Strong~\cite{frome2013devise,radford2021learning} \& partial~\cite{vendrov2015order,zhang2016learning} coordination \\
\Xcline{2-5}{0.5\arrayrulewidth}
& \multirow{1}{*}{Fission (\S\cref{sec:representation3})} & Modality-level~\cite{hessel2020emap,tsai2018learning} \& fine-grained~\cite{abavisani2018deep,chen2021multimodal} fission \\
\hline \hline

\multirow{3}{*}{Alignment (\S\cref{sec:alignment})} & \multirow{1}{*}{Discrete connections (\S\cref{sec:alignment1})} & Local~\cite{cirik2018visual,hsu2018unsupervised} \& global~\cite{li2022clip} alignment \\
\Xcline{2-5}{0.5\arrayrulewidth}
& \multirow{1}{*}{Continuous alignment (\S\cref{sec:alignment2})} & Warping~\cite{hu2019multimodal,haresh2021learning} \& segmentation~\cite{sun2019videobert} \\
\Xcline{2-5}{0.5\arrayrulewidth}
& \multirow{1}{*}{Contextualization (\S\cref{sec:alignment3})} & Joint~\cite{li2019visualbert}, cross-modal~\cite{hendricks2021decoupling,lu2019vilbert} \& graphical~\cite{yang2021mtag} &  \\
\hline \hline

\multirow{4}{*}{Reasoning (\S\cref{sec:reasoning})} & \multirow{1}{*}{Structure modeling (\S\cref{sec:reasoning1})} & Hierarchical~\cite{andreas2016neural}, temporal~\cite{xiong2016dynamic}, interactive~\cite{luketina2019survey} \& discovery~\cite{perez2019mfas} \\
\Xcline{2-5}{0.5\arrayrulewidth}
& \multirow{1}{*}{Intermediate concepts (\S\cref{sec:reasoning2})} & Attention~\cite{xu2015show}, discrete symbols~\cite{amizadeh2020neuro,vedantam2019probabilistic} \& language~\cite{hudson2019learning,zeng2022socratic} \\
\Xcline{2-5}{0.5\arrayrulewidth}
& \multirow{1}{*}{Inference paradigm (\S\cref{sec:reasoning3})} & Logical~\cite{gokhale2020vqa,suzuki2019multimodal} \& causal~\cite{agarwal2020towards,niu2021counterfactual,yi2019clevrer} \\
\Xcline{2-5}{0.5\arrayrulewidth}
& \multirow{1}{*}{External knowledge (\S\cref{sec:reasoning4})} & Knowledge graphs~\cite{gui2021kat,zhu2015building} \& commonsense~\cite{park2020visualcomet,zellers2019vcr} \\
\hline \hline

\multirow{3}{*}{Generation (\S\cref{sec:generation})} & \multirow{1}{*}{Summarization (\S\cref{sec:generation1})} & Extractive~\cite{chen2018extractive,uzzaman2011multimodal} \& abstractive~\cite{li2017multi,palaskar2019multimodal} \\
\Xcline{2-5}{0.5\arrayrulewidth}
& \multirow{1}{*}{Translation (\S\cref{sec:generation2})} & Exemplar-based~\cite{karpathy2014deep,lebret2015phrase} \& generative~\cite{ahuja2020style,jamaludin2019you,ramesh2021zero} \\
\Xcline{2-5}{0.5\arrayrulewidth}
& \multirow{1}{*}{Creation (\S\cref{sec:generation3})} & Conditional decoding~\cite{denton2018stochastic,oord2018parallel,zhu2021arbitrary} \\
\Xcline{2-5}{0.5\arrayrulewidth}
\hline \hline

\multirow{3}{*}{Transference (\S\cref{sec:transference})} & \multirow{1}{*}{Cross-modal transfer (\S\cref{sec:transference1})} & Tuning~\cite{rahman2020integrating,tsimpoukelli2021multimodal}, multitask~\cite{singh2021flava,liang2022highmmt} \& transfer~\cite{lu2021pretrained} \\
\Xcline{2-5}{0.5\arrayrulewidth}
& \multirow{1}{*}{Co-learning (\S\cref{sec:transference2})} & Representation~\cite{jia2021scaling,zadeh2020foundations} \& generation~\cite{pham2019found,tan2020vokenization} \\
\Xcline{2-5}{0.5\arrayrulewidth}
& \multirow{1}{*}{Model Induction (\S\cref{sec:transference3})} & Co-training~\cite{blum1998combining,dunnmon2020cross} \& co-regularization~\cite{sridharan2008information,yang2019comprehensive} &  \\
\hline \hline

\multirow{3}{*}{Quantification (\S\cref{sec:quantification})} & \multirow{1}{*}{Heterogenity (\S\cref{sec:quantification1})} & Importance~\cite{gat2021perceptual,park2018multimodal}, bias~\cite{hendricks2018women,pena2020faircvtest} \& noise~\cite{ma2021smil} \\
\Xcline{2-5}{0.5\arrayrulewidth}
& \multirow{1}{*}{Interconnections (\S\cref{sec:quantification2})} & Connections~\cite{aflalo2022vl,cao2020behind,thrush2022winoground} \& interactions~\cite{hessel2020emap,liang2022multiviz,wang2021m2lens} \\
\Xcline{2-5}{0.5\arrayrulewidth}
& \multirow{1}{*}{Learning (\S\cref{sec:quantification3})} & Generalization~\cite{liang2022highmmt,reed2022generalist}, optimization~\cite{wang2020makes,wu2022characterizing} \& tradeoffs~\cite{liang2021multibench} \\
\hline \hline
\end{tabular}
\vspace{-4mm}
\label{table:all}
\end{table*}

Building on these three core principles and on our detailed review of recent work, we propose a new taxonomy to characterize the core technical challenges in multimodal research: representation, alignment, reasoning, generation, transference, and quantification. In Table~\cref{table:all} we summarize our full taxonomy of these six core challenges, their subchallenges, categories of corresponding approaches, and recent examples in each category. In the following sections, we describe our new taxonomy in detail and also revisit the principles of heterogeneity, connections, and interactions to see how they pose research questions and inspire research in each of these six challenges.  

\vspace{-2mm}
\section{Challenge 1: Representation}
\label{sec:representation}
\vspace{-1mm}

\begin{figure}[t]
\centering
\includegraphics[width=0.7\linewidth]{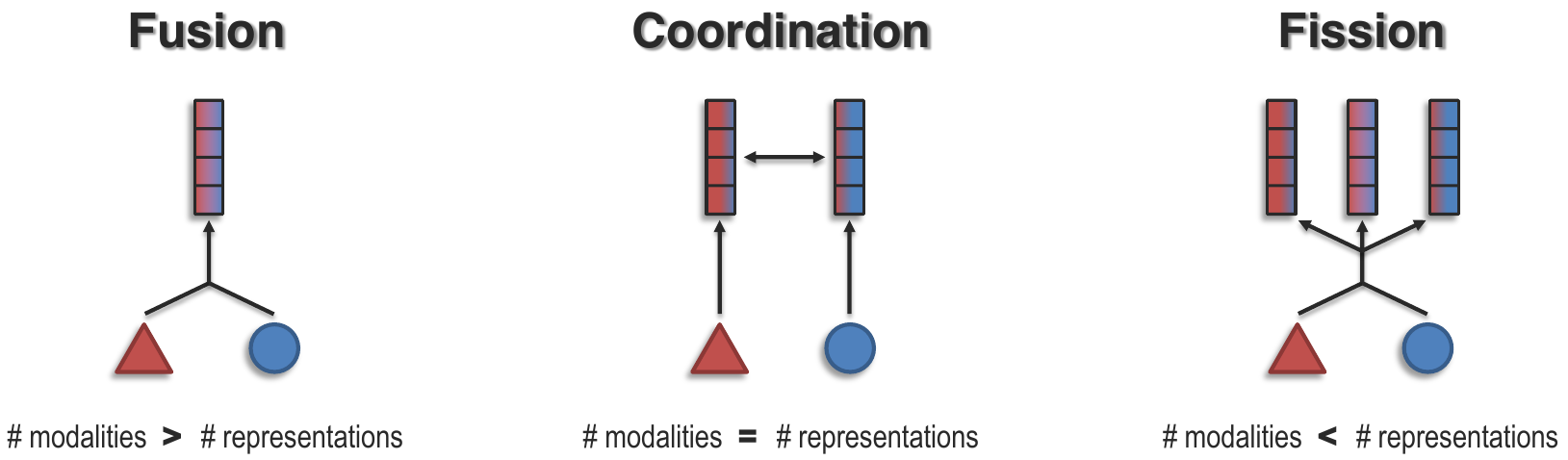}
\vspace{-2mm}
\caption{Challenge 1 aims to learn \textbf{representations} that reflect cross-modal interactions between individual modality elements, through (1) \textit{fusion}: integrating information to reduce the number of separate representations, (2) \textit{coordination}: interchanging cross-modal information with the goal of keeping the same number of representations but improving multimodal contextualization, and (3) \textit{fission}: creating a larger set of decoupled representations that reflects knowledge about internal structure.}
\label{fig:rep}
\vspace{-4mm}
\end{figure}

The first fundamental challenge is to learn representations that reflect cross-modal interactions between individual elements across different modalities. This challenge can be seen as learning a `local' representation between elements, or a representation using holistic features.
This section covers (1) \textit{representation fusion}: integrating information from 2 or more modalities, effectively reducing the number of separate representations, (2) \textit{representation coordination}: interchanging cross-modal information with the goal of keeping the same number of representations but improving multimodal contextualization, and (3) \textit{representation fission}: creating a new decoupled set of representations, usually larger number than the input set, that reflects knowledge about internal structure such as data clustering or factorization (Figure~\cref{fig:rep}).

\vspace{-2mm}
\subsection{Subchallenge 1a: Representation Fusion}
\label{sec:representation1}
\vspace{-1mm}

\begin{figure}[t]
\centering
\vspace{-0mm}
\includegraphics[width=0.6\linewidth]{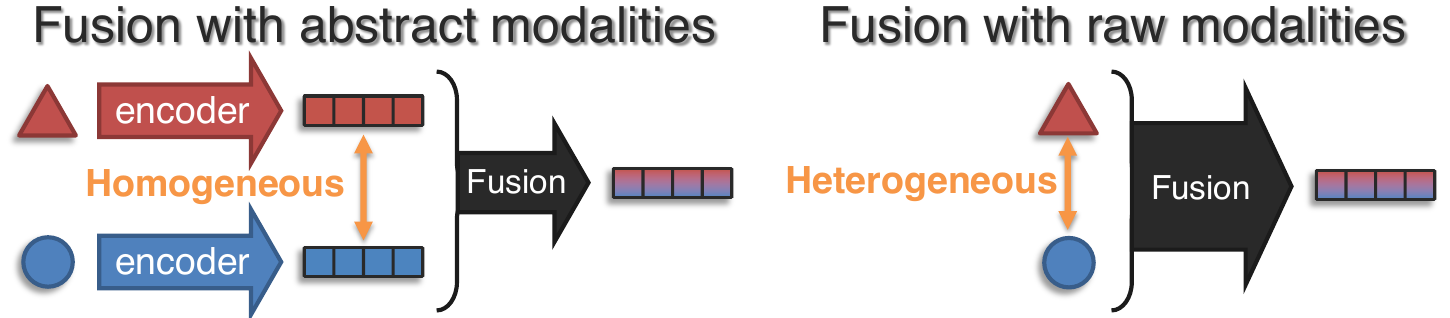}
\vspace{-2mm}
\caption{We categorize \textbf{representation fusion} approaches into (1) \textit{fusion with abstract modalities}, where unimodal encoders first capture a holistic representation of each element before fusion at relatively homogeneous representations, and (2) \textit{fusion with raw modalities} which entails representation fusion at very early stages, perhaps directly involving heterogeneous raw modalities.}
\label{fig:rep1}
\vspace{-4mm}
\end{figure}

Representation fusion aims to learn a joint representation that models cross-modal interactions between individual elements of different modalities, effectively \textit{reducing} the number of separate representations. We categorize these approaches into \textit{fusion with abstract modalities} and \textit{fusion with raw modalities} (Figure~\cref{fig:rep1}). In fusion with abstract modalities, suitable unimodal encoders are first applied to capture a holistic representation of each element (or modality entirely), after which several building blocks for representation fusion are used to learn a joint representation. As a result, fusion happens at the abstract representation level. On the other hand, fusion with raw modalities entails representation fusion at very early stages with minimal preprocessing, perhaps even involving raw modalities themselves.

\textbf{Fusion with abstract modalities}: We begin our treatment of representation fusion of abstract representations with \textit{additive and multiplicative interactions}. These operators can be seen as differentiable building blocks combining information from two streams of data that can be flexibly inserted into almost any unimodal machine learning pipeline. Given unimodal data or features $\mathbf{x}_1$ and $\mathbf{x}_2$, additive fusion can be seen as learning a new joint representation $\mathbf{z}_\textrm{mm} = w_0 + w_1 \mathbf{x}_1 + w_2 \mathbf{x}_2 + \epsilon$, where $w_1$ and $w_2$ are the weights learned for additive fusion of $\mathbf{x}_1$ and $\mathbf{x}_2 $, $w_0$ the bias term, and $\epsilon$ the error term. If the joint representation $\mathbf{z}_\textrm{mm}$ is directly taken as a prediction $\hat{y}$, then additive fusion resembles late or ensemble fusion $\hat{y} = f_1(\mathbf{x}_1) + f_2(\mathbf{x}_2)$ with unimodal predictors $f_1$ and $f_2$~\citep{friedman2008predictive}. Otherwise, the additive representation $\mathbf{z}_\textrm{mm}$ can also undergo subsequent unimodal or multimodal processing~\citep{baltruvsaitis2018multimodal}.
Multiplicative interactions extend additive interactions to include a cross term $w_3 (\mathbf{x}_1 \times \mathbf{x}_2)$. These models have been used extensively in statistics, where it can be interpreted as a \textit{moderation} effect of $\mathbf{x}_1$ affecting the linear relationship between $\mathbf{x}_2$ and $y$~\cite{baron1986moderator}. Overall, purely additive interactions $\mathbf{z}_\textrm{mm} = w_0 + w_1 \mathbf{x}_1 + w_2 \mathbf{x}_2$ can be seen as a first-order polynomial between input modalities $\mathbf{x}_1$ and $\mathbf{x}_2$, combining additive and multiplicative $\mathbf{z}_\textrm{mm} = w_0 + w_1 \mathbf{x}_1 + w_2 \mathbf{x}_2 + w_3 (\mathbf{x}_1 \times \mathbf{x}_2)$ captures a second-order polynomial.

To further go beyond first and second-order interactions, \textit{tensors} are specifically designed to explicitly capture higher-order interactions across modalities~\cite{zadeh2017tensor}. Given unimodal data $\mathbf{x}_1, \mathbf{x}_2$, tensors are defined as $\mathbf{z}_\textrm{mm} = \mathbf{x}_{1}\otimes \mathbf{x}_{2}$ where $\otimes$ denotes an outer product~\citep{ben2017mutan,fukui2016multimodal}. Tensor products of higher order represent polynomial interactions of higher order between elements~\citep{hou2019deep}.
However, computing tensor products is expensive since their dimension scales exponentially with the number of modalities, so several efficient approximations based on low-rank decomposition have been proposed~\cite{hou2019deep,liu2018efficient}. Finally, \textit{Multiplicative Interactions (MI)} generalize additive and multiplicative operators to include learnable parameters that capture second-order interactions~\cite{jayakumar2020multiplicative}. In its most general form, MI defines a bilinear product $\mathbf{z}_\textrm{mm} = \mathbf{x}_1 \mathbb{W} \mathbf{x}_2 + \mathbf{x}_1^\top \mathbf{U} + \mathbf{V} \mathbf{x}_2 + \mathbf{b}$ where $\mathbb{W}, \mathbf{U}, \mathbf{Z}$, and $\mathbf{b}$ are trainable parameters.

\textit{Multimodal gated units/attention units} learn representations that dynamically change for every input~\cite{chaplot2017gated,wang2020makes}. Its general form can be written as $\mathbf{z}_\textrm{mm} = \mathbf{x}_1 \odot h(\mathbf{x}_2)$, where $h$ represents a function with sigmoid activation and $\odot$ denotes element-wise product. $h(\mathbf{x}_2)$ is commonly referred to as `attention weights' learned from $\mathbf{x}_2$ to attend on $\mathbf{x}_1$. Recent work has explored more expressive forms of learning attention weights such as using Query-Key-Value mechanisms~\cite{tsai2019multimodal}, fully-connected neural network layers~\citep{arevalo2017gated,chaplot2017gated}, or even hard gated units for sharper attention~\citep{chen2017multimodal}.

\textbf{Fusion with raw modalities} entails representation fusion at very early stages, perhaps even involving raw modalities themselves. These approaches typically bear resemblance to early fusion~\citep{baltruvsaitis2018multimodal}, which performs concatenation of input data before applying a prediction model (i.e., $\mathbf{z}_\textrm{mm} = \left[ \mathbf{x}_1, \mathbf{x}_2\right]$). Fusing at the raw modality level is more challenging since raw modalities are likely to exhibit more dimensions of heterogeneity. Nevertheless,~\citet{barnum2020benefits} demonstrated robustness benefits of fusion at early stages, while~\citet{gadzicki2020early} also found that complex early fusion can outperform abstract fusion. To account for the greater heterogeneity during complex early fusion, many approaches rely on generic encoders that are applicable to both modalities, such as convolutional layers~\citep{barnum2020benefits,gadzicki2020early} and Transformers~\citep{liang2022highmmt,likhosherstov2022polyvit}. However, do these complex non-additive fusion models actually learn non-additive interactions between modality elements? Not necessarily, according to~\citet{hessel2020emap}. We cover these fundamental analysis questions and more in the quantification challenge (\S\cref{sec:quantification}).

\vspace{-2mm}
\subsection{Subchallenge 1b: Representation Coordination}
\label{sec:representation2}
\vspace{-1mm}

\begin{figure}[t]
\centering
\vspace{-0mm}
\includegraphics[width=0.6\linewidth]{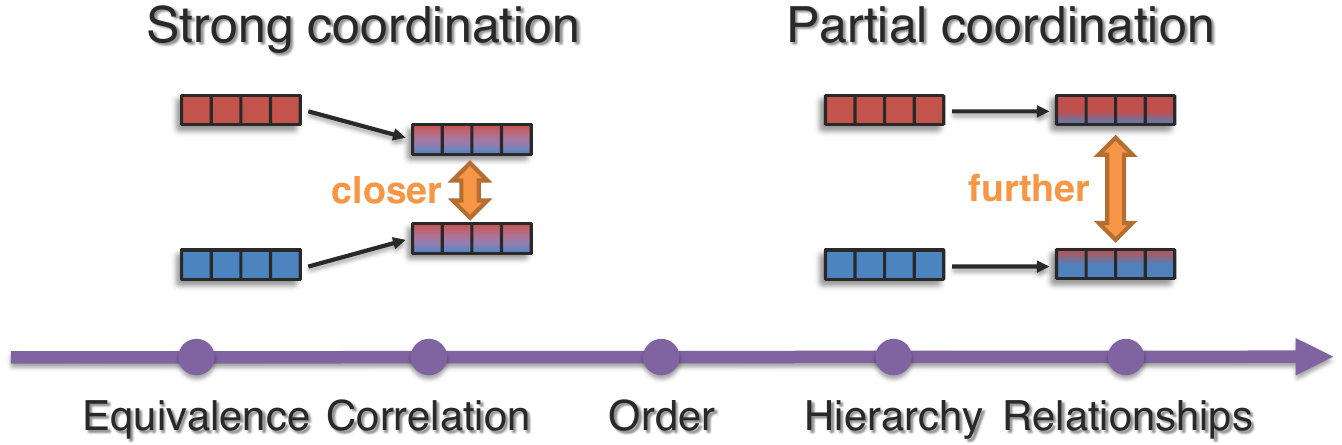}
\vspace{-2mm}
\caption{There is a spectrum of \textbf{representation coordination} functions: \textit{strong coordination} aims to enforce strong equivalence in all dimensions, whereas in \textit{partial coordination} only certain dimensions may be coordinated to capture more general connections such as correlation, order, hierarchies, or relationships.}
\label{fig:rep2}
\vspace{-4mm}
\end{figure}

Representation coordination aims to learn multimodal contextualized representations that are coordinated through their interconnections (Figure~\cref{fig:rep2}). In contrast to representation fusion, coordination keeps the same number of representations but improves multimodal contextualization. We start our discussion with \textit{strong coordination} that enforces strong equivalence between modality elements, before moving on to \textit{partial coordination} that captures more general connections such as correlation, order, hierarchies, or relationships beyond similarity.

\textbf{Strong coordination} aims to bring semantically corresponding modalities close together in a coordinated space, thereby enforcing strong \textit{equivalence} between modality elements. For example, these models would encourage the representation of the word `dog' and an image of a dog to be close (i.e., semantically positive pairs), while the distance between the word `dog' and an image of a car to be far apart (i.e., semantically negative pairs)~\cite{frome2013devise}. The coordination distance is typically cosine distance~\cite{mekhaldi2007multimodal,wehrmann2018order} or max-margin losses~\cite{hu2019deep}. Recent work has explored large-scale representation coordination by scaling up contrastive learning of image and text pairs~\cite{radford2021learning}, and also found that contrastive learning provably captures redundant information across the two views~\citep{tian2020contrastive,tosh2021contrastive} (but not non-redundant information). In addition to contrastive learning, several approaches instead learn a coordinated space by mapping corresponding data from one modality to another~\cite{dyer2014notes}. For example,~\citet{socher2013zero} maps image embeddings into word embedding spaces for zero-shot image classification. Similar ideas were used to learn coordinated representations between text, video, and audio~\cite{pham2019found}, as well as between pretrained language models and image features~\cite{tan2020vokenization}.

\textbf{Partial coordination}: Instead of strictly capturing equivalence via strong coordination, partial coordination instead captures more general modality connections such as correlation, order, hierarchies, or relationships. To achieve these goals, partially coordinated models enforce different types of constraints on the representation space beyond semantic similarity, and perhaps only on certain dimensions of the representation.

\textit{Canonical correlation analysis} (CCA) computes a linear projection that maximizes the correlation between two random variables while enforcing each dimension in a new representation to be orthogonal to each other~\citep{thompson2000canonical}.
CCA models have been used extensively for cross-modal retrieval~\cite{Rasiwasia2010} audio-visual signal analysis~\cite{Sargin2007}, and emotion recognition~\cite{nemati2019hybrid}. To increase the expressiveness of CCA, several nonlinear extensions have been proposed including Kernel CCA~\cite{lai2000kernel}, Deep CCA~\cite{andrew2013deep}, and CCA Autoencoders~\cite{wang2015deep}.

\textit{Ordered and hierarchical spaces}: Another example of representation coordination comes from order-embeddings of images and language~\cite{vendrov2015order}, which aims to capture a partial order on the language and image embeddings to enforce a hierarchy in the coordinated space. A similar model using denotation graphs was also proposed by~\citet{Young2014} where denotation graphs are used to induce such a partial ordering hierarchy.

\textit{Relationship coordination}: In order to learn a coordinated space that captures semantic relationships between elements beyond correspondences,~\citet{zhang2016learning} use structured representations of text and images to create multimodal concept taxonomies.~\citet{delaherche2010multimodal} learn coordinated representations capturing hierarchical relationships, while~\citet{alviar2020multimodal} apply multiscale coordination of speech and music using partial correlation measures. Finally,~\citet{xu2015multi} learn coordinated representations using a Cauchy loss to strengthen robustness to outliers.

\vspace{-2mm}
\subsection{Subchallenge 1c: Representation Fission}
\label{sec:representation3}
\vspace{-1mm}

\begin{figure}[t]
\centering
\vspace{-1mm}
\includegraphics[width=0.5\linewidth]{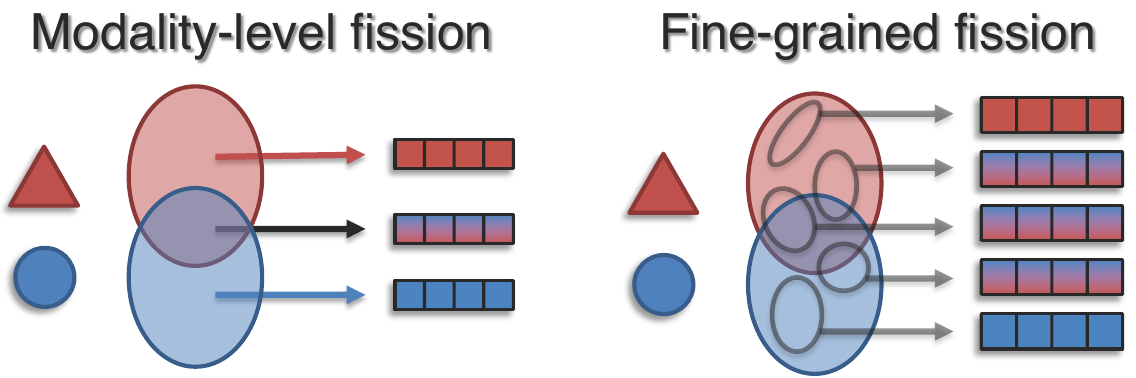}
\vspace{-3mm}
\caption{\textbf{Representation fission} creates a larger set of decoupled representations that reflects knowledge about internal structure. (1) \textit{Modality-level fission} factorizes into modality-specific information primarily in each modality, and multimodal information redundant in both modalities, while (2) \textit{fine-grained fission} attempts to further break multimodal data down into individual subspaces.}
\label{fig:rep3}
\vspace{-4mm}
\end{figure}

Finally, representation fission aims to create a new decoupled set of representations (usually a larger number than the input representation set) that reflects knowledge about internal multimodal structure such as data clustering, independent factors of variation, or modality-specific information. In comparison with joint and coordinated representations, representation fission enables careful interpretation and fine-grained controllability. Depending on the granularity of decoupled factors, methods can be categorized into \textit{modality-level} and \textit{fine-grained} fission (Figure~\cref{fig:rep3}).

\textbf{Modality-level fission} aims to factorize into modality-specific information primarily in each modality and multimodal information redundant in both modalities~\cite{hsu2018disentangling,tsai2018learning}. \textit{Disentangled representation learning} aims to learn mutually independent latent variables that each explain a particular variation of the data~\cite{Bengio:2013:RLR:2498740.2498889,higgins2016beta}, and has been useful for modality-level fission by enforcing independence constraints on modality-specific and multimodal latent variables~\cite{hsu2018disentangling,tsai2018learning}.~\citet{tsai2018learning} and~\citet{hsu2018disentangling} study factorized multimodal representations and demonstrate the importance of modality-specific and multimodal factors towards generation and prediction.~\citet{shi2019variational} study modality-level fission in multimodal variational autoencoders using a mixture-of-experts layer, while~\citet{wu2018multimodal} instead use a product-of-experts layer.

\textit{Post-hoc representation disentanglement} is suitable when it is difficult to retrain a disentangled model, especially for large pretrained multimodal models. Empirical multimodally-additive function projection (EMAP)~\cite{hessel2020emap} is an approach for post-hoc disentanglement of the effects of unimodal (additive) contributions from cross-modal interactions in multimodal tasks, which works for arbitrary multimodal models and tasks. EMAP is also closely related to the use of Shapley values for feature disentanglement and interpretation~\cite{merrick2020explanation}, which can also be used for post-hoc representation disentanglement in general models.

\textbf{Fine-grained fission}: Beyond factorizing only into individual modality representations, fine-grained fission attempts to further break multimodal data down into the individual subspaces covered by the modalities~\cite{vidal2011subspace}. \textit{Clustering} approaches that group data based on semantic similarity~\cite{madhulatha2012overview} have been integrated with multimodal networks for end-to-end representation fission and prediction. For example,~\citet{hu2019deep} combine $k$-means clustering in representations with unsupervised audiovisual learning.~\citet{chen2021multimodal} combine $k$-means clustering with self-supervised contrastive learning on videos.
Subspace clustering~\cite{abavisani2018deep}, approximate graph Laplacians~\cite{khan2019approximate}, conjugate mixture models~\cite{khalidov2011conjugate}, and dictionary learning~\cite{kim2016joint} have also been integrated with multimodal models. Motivated by similar goals of representation fission, \textit{matrix factorization} techniques have also seen several applications in multimodal prediction~\cite{aktukmak2019probabilistic} and image retrieval~\cite{caicedo2012online}.
\vspace{-2mm}
\section{Challenge 2: Alignment}
\label{sec:alignment}
\vspace{-1mm}

\begin{figure}[t]
\centering
\vspace{-0mm}
\includegraphics[width=0.7\linewidth]{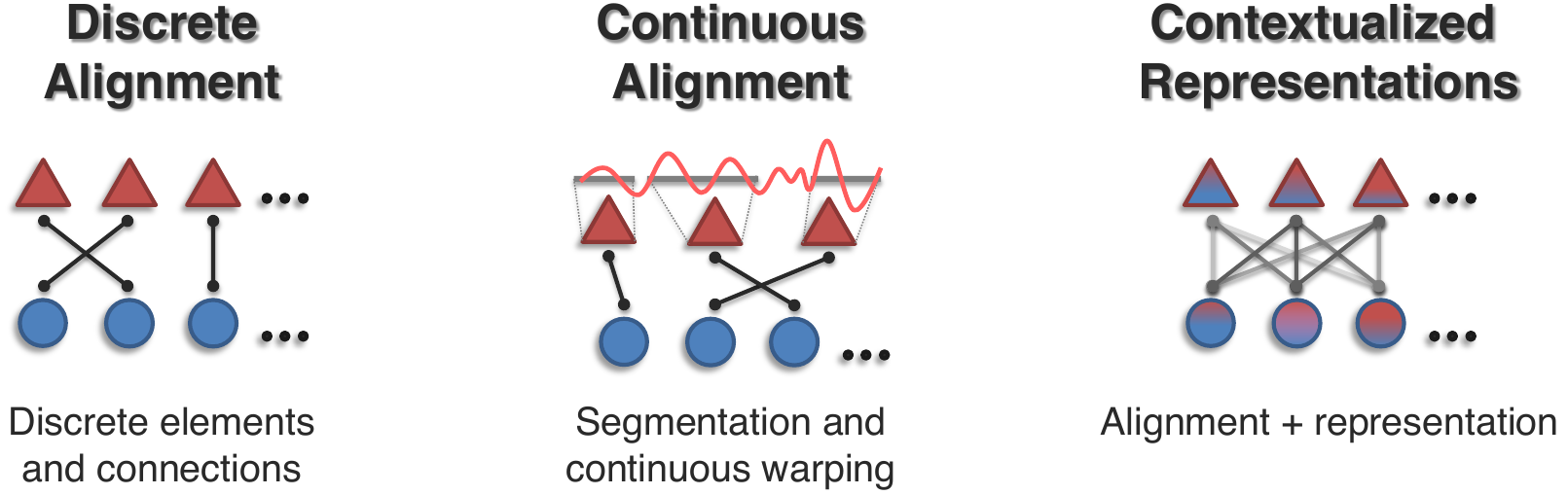}
\vspace{-2mm}
\caption{\textbf{Alignment} aims to identify cross-modal connections and interactions between modality elements. Recent work has involved (1) \textit{discrete alignment} to identify connections among discrete elements, (2) \textit{continuous alignment} of continuous signals with ambiguous segmentation, and (3) \textit{contextualized representation} learning to capture these cross-modal interactions between connected elements.}
\label{fig:align}
\vspace{-4mm}
\end{figure}

A second challenge is to identify cross-modal connections and interactions between elements of multiple modalities. For example, when analyzing the speech and gestures of a human subject, how can we align specific gestures with spoken words or utterances?
Alignment between modalities is challenging since it may depend on long-range dependencies, involves ambiguous segmentation (e.g., words or utterances), and could be either one-to-one, many-to-many, or not exist at all.
This section covers recent work in multimodal alignment involving (1) \textit{discrete alignment}: identifying connections between discrete elements across modalities, (2) \textit{continuous alignment}: modeling alignment between continuous modality signals with ambiguous segmentation, and (3) \textit{contextualized representations}: learning better multimodal representations by capturing cross-modal interactions between elements (Figure~\cref{fig:align}).

\vspace{-2mm}
\subsection{Subchallenge 2a: Discrete Alignment}
\label{sec:alignment1}
\vspace{-1mm}

\begin{figure}[t]
\centering
\vspace{-0mm}
\includegraphics[width=0.4\linewidth]{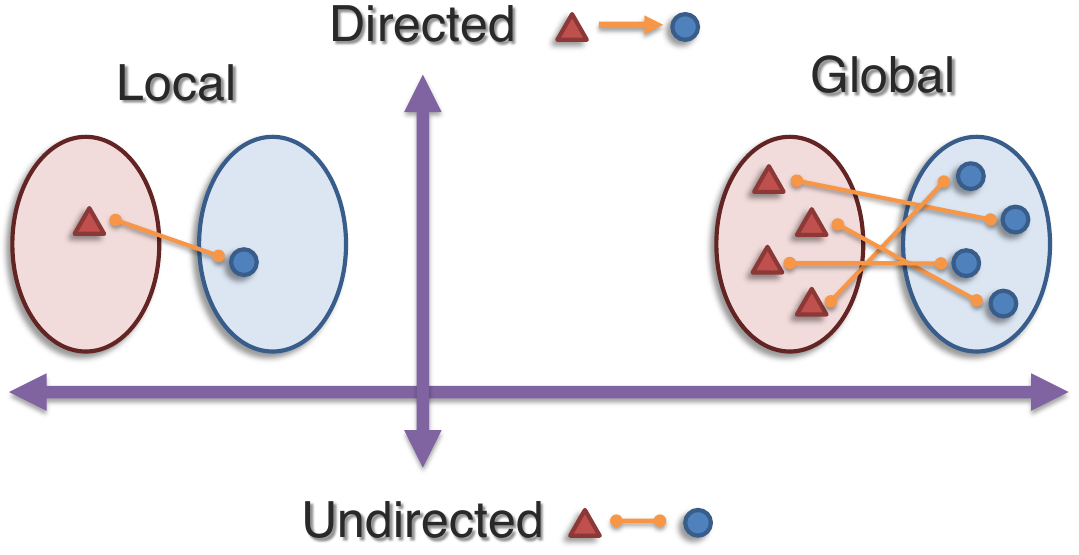}
\vspace{-2mm}
\caption{\textbf{Discrete alignment} identifies connections between discrete elements, spanning (1) \textit{local alignment} to discover connections given matching pairs, and (2) \textit{global alignment} where alignment must be performed globally to learn both the connections and matchings between modality elements.}
\label{fig:align1}
\vspace{-4mm}
\end{figure}

The first subchallenge aims to identify connections between discrete elements of multiple modalities. We describe recent work in (1) \textit{local alignment} to discover connections between a given matching pair of modality elements, and (2) \textit{global alignment} where alignment must be performed globally to learn both the connections and matchings (Figure~\cref{fig:align1}).

\textbf{Local alignment} between connected elements is particularly suitable for multimodal tasks where there is clear segmentation into discrete elements such as words in text or object bounding boxes in images or videos (e.g., tasks such as visual coreference resolution~\cite{kottur2018visual}, visual referring expression recognition~\cite{cirik2018using,cirik2020refer360}, and cross-modal retrieval~\cite{frome2013devise,plummer2015flickr30k}). When we have supervised data in the form of connected modality pairs, \textit{contrastive learning} is a popular approach where the goal is to match representations of the same concept expressed in different modalities~\cite{baltruvsaitis2018multimodal}. Several objective functions for learning aligned spaces from varying quantities of paired~\cite{cao2017transitive,huang2017cross} and unpaired~\cite{grave2019unsupervised} data have been proposed. Many of the ideas that enforce strong~\cite{frome2013devise,liang2021cross} or partial~\citep{andrew2013deep,vendrov2015order,zhang2016learning} representation coordination (\S\cref{sec:representation2}) are also applicable for local alignment. Several examples include aligning books with their corresponding movies/scripts~\cite{zhu2015aligning}, matching referring expressions to visual objects~\cite{mao2016generation}, and finding similarities between image regions and their descriptions~\cite{hu2016natural}. Methods for local alignment have also enabled the learning of shared semantic concepts not purely based on language but also on additional modalities such as vision~\cite{huang2017cross}, sound~\cite{cirik2018visual,socher2013zero}, and multimedia~\cite{zhu2015aligning} that are useful for downstream tasks.

\textbf{Global alignment}: When the ground-truth modality pairings are not available, alignment must be performed globally between all elements across both modalities. Optimal transport (OT)-based approaches~\citep{villani2009optimal} (which belong to a broader set of matching algorithms) are a potential solution since they jointly optimize the coordination function and optimal coupling between modality elements by posing alignment as a divergence minimization problem. These approaches are useful for aligning multimodal representation spaces~\cite{pramanick2022multimodal,li2022clip}. To alleviate computational issues, several recent advances have integrated them with neural networks~\cite{chen2020graph}, approximated optimal transport with entropy regularization~\cite{wei2018unsupervised}, and formulated convex relaxations for efficient learning~\cite{grave2019unsupervised}.

\vspace{-2mm}
\subsection{Subchallenge 2b: Continuous Alignment}
\label{sec:alignment2}
\vspace{-1mm}

So far, one important assumption we have made is that modality elements are already segmented and discretized. While certain modalities display clear segmentation (e.g., words/phrases in a sentence or object regions in an image), there are many cases where the segmentation is not readily provided, such as in continuous signals (e.g, financial or medical time-series), spatio-temporal data (e.g., satellite or weather images), or data without clear semantic boundaries (e.g., MRI images). In these settings, methods based on warping and segmentation have been recently proposed:

\textbf{Continuous warping} aims to align two sets of modality elements by representing them as continuous representation spaces and forming a bridge between these representation spaces. \textit{Adversarial training} is a popular approach to warp one representation space into another. Initially used in domain adaptation~\cite{ben2006analysis}, adversarial training learns a domain-invariant representation across domains where a domain classifier is unable to identify which domain a feature came from~\cite{ajakan2014domain}. These ideas have been extended to align multimodal spaces~\cite{hsu2018unsupervised,hu2019multimodal,munro2020multi}.~\citet{hsu2018unsupervised} use adversarial training to align images and medical reports,~\citet{hu2019multimodal} design an adversarial network for cross-modal retrieval, and~\citet{munro2020multi} design both self-supervised alignment and adversarial alignment objectives for multimodal action recognition. \textit{Dynamic time warping (DTW)}~\cite{kruskal1983overview} is a related approach to segment and align multi-view time series data. DTW measures the similarity between two sequences and finds an optimal match between them by time warping (inserting frames) such that they are aligned across segmented time boundaries. For multimodal tasks, it is necessary to design similarity metrics between modalities~\cite{anguera2014audio,tapaswi2015book2movie}. DTW was extended using CCA to map the modalities to a coordinated space, allowing for both alignment (through DTW) and coordination (through CCA) between different modality streams jointly~\cite{trigeorgis2017deep}.

\begin{figure}[t]
\centering
\vspace{-1mm}
\includegraphics[width=0.4\linewidth]{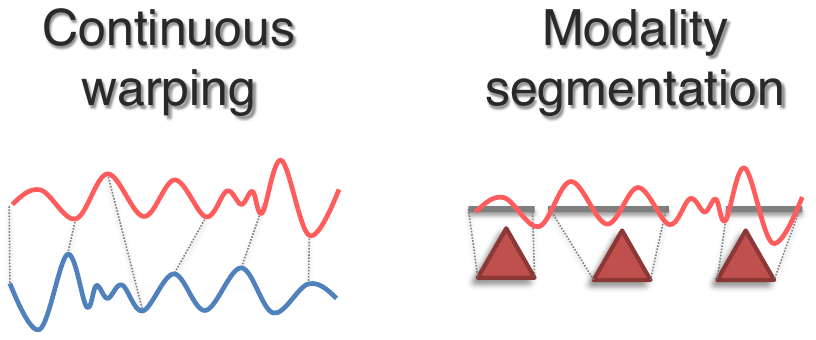}
\vspace{-3mm}
\caption{\textbf{Continuous alignment} tackles the difficulty of aligning continuous signals where element segmentation is not readily available. We cover related work in (1) \textit{continuous warping} of representation spaces and (2) \textit{modality segmentation} of continuous signals into discrete elements at an appropriate granularity.}
\label{fig:align2}
\vspace{-4mm}
\end{figure}

\textbf{Modality segmentation} involves dividing high-dimensional data into elements with semantically-meaningful boundaries. A common problem involves \textit{temporal segmentation}, where the goal is to discover the temporal boundaries across sequential data. Several approaches for temporal segmentation include forced alignment, a popular approach to align discrete speech units with individual words in a transcript~\cite{yuan2008speaker}.~\citet{malmaud2015s} explore multimodal alignment using a factored hidden Markov model to align ASR transcripts to the ground truth. \textit{Clustering} approaches have also been used to group continuous data based on semantic similarity~\cite{madhulatha2012overview}. Clustering-based discretization has recently emerged as an important preprocessing step for generalizing language-based pretraining (with clear word/bytepair segmentation boundaries and discrete elements) to video or audio-based pretraining (without clear segmentation boundaries and continuous elements). By clustering raw video or audio features into a discrete set, approaches such as VideoBERT~\citep{sun2019videobert} perform masked pretraining on raw video and audio data. Similarly, approaches such as DALL.E~\citep{ramesh2021zero}, VQ-VAE~\citep{van2017neural}, and CMCM~\citep{liu2022cross} also utilize discretized intermediate layers obtained via vector quantization and showed benefits in modality alignment.

\vspace{-2mm}
\subsection{Subchallenge 2c: Contextualized Representations}
\label{sec:alignment3}
\vspace{-1mm}

\begin{figure}[t]
\centering
\vspace{-0mm}
\includegraphics[width=0.6\linewidth]{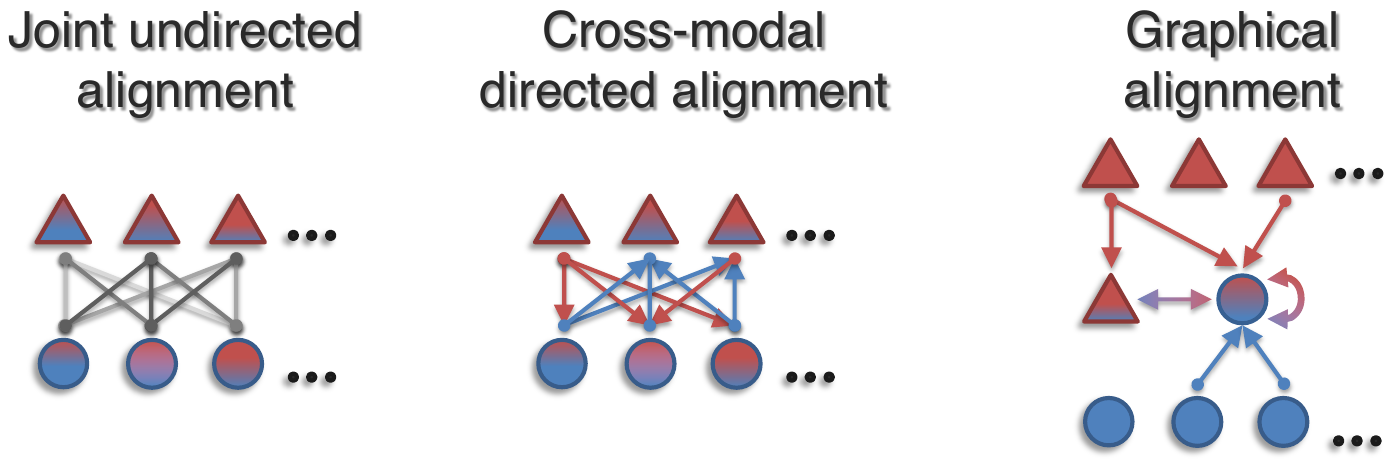}
\vspace{-2mm}
\caption{\textbf{Contextualized representation} learning aims to model modality connections to learn better representations. Recent directions include (1) \textit{joint undirected alignment} that captures undirected symmetric connections, (2) \textit{cross-modal directed alignment} that models asymmetric connections in a directed manner, and (3) \textit{graphical alignment} that generalizes the sequential pattern into arbitrary graph structures.}
\label{fig:align3}
\vspace{-4mm}
\end{figure}

Finally, contextualized representation learning aims to model all modality connections and interactions to learn better representations. Contextualized representations have been used as an intermediate (often latent) step enabling better performance on a number of downstream tasks including speech recognition, machine translation, media description, and visual question-answering. We categorize work in contextualized representations into (1) \textit{joint undirected alignment}, (2) \textit{cross-modal directed alignment}, and (3) \textit{alignment with graph networks} (Figure~\cref{fig:align3}).

\textbf{Joint undirected alignment} aims to capture undirected connections across pairs of modalities, where the connections are symmetric in either direction. This is commonly referred to in the literature as unimodal, bimodal, trimodal interactions, and so on~\cite{macaluso2005multisensory}. Joint undirected alignment is typically captured by parameterizing models with alignment layers and training end-to-end for a multimodal task. These alignment layers can include attention weights~\cite{chaplot2017gated}, tensor products~\cite{liu2018efficient,zadeh2017tensor}, and multiplicative interactions~\cite{jayakumar2020multiplicative}. More recently, transformer models~\cite{vaswani2017attention} have emerged as powerful encoders for sequential data by automatically aligning and capturing complementary features at different time steps. Building upon the initial text-based transformer model, multimodal transformers have been proposed that perform joint alignment using a full self-attention over modality elements concatenated across the sequence dimension (i.e., early fusion)~\cite{li2019visualbert,sun2019videobert}. As a result, all modality elements become jointly connected to all other modality elements similarly (i.e., modeling all connections using dot-product similarity kernels).

\textbf{Cross-modal directed alignment} relates elements from a source modality in a directed manner to a target modality, which can model asymmetric connections. For example, \textit{temporal attention models} use alignment as a latent step to improve many sequence-based tasks~\cite{xiong2016dynamic,zeng2017leveraging}. These attention mechanisms are typically directed from the output to the input so that the resulting weights reflect a soft alignment distribution over the input. \textit{Multimodal transformers} perform directed alignment using query-key-value attention mechanisms to attend from one modality's sequence to another, before repeating in a bidirectional manner. This results in two sets of asymmetric contextualized representations to account for the possibly asymmetric connections between modalities~\cite{lu2019vilbert,tan2019lxmert,tsai2019multimodal}. 
These methods are useful for sequential data by automatically aligning and capturing complementary features at different time-steps~\cite{tsai2019multimodal}. 
Self-supervised multimodal pretraining has also emerged as an effective way to train these architectures, with the aim of learning general-purpose representations from larger-scale unlabeled multimodal data before transferring to specific downstream tasks via supervised fine-tuning~\cite{li2019visualbert}. These pretraining objectives typically consist of unimodal masked prediction, crossmodal masked prediction, and multimodal alignment prediction~\cite{hendricks2021decoupling}.

\textbf{Graphical alignment} generalizes the sequential pattern seen in undirected or directed alignment into arbitrary graph structures between elements. This has several benefits since it does not require all elements to be connected, and allows the user to choose different edge functions for different connections. Solutions in this subcategory typically make use of graph neural networks~\citep{velivckovic2018graph} to recursively learn element representations contextualized with the elements in locally connected neighborhoods~\citep{scarselli2008graph,velivckovic2018graph}. These approaches have been applied for multimodal sequential data through MTAG~\citep{yang2021mtag} that captures connections in human videos, and F2F-CL~\citep{wilf2022face} that additionally adds factorizes nodes along speaker turns.
\begin{figure}[t]
\centering
\vspace{-0mm}
\includegraphics[width=0.6\linewidth]{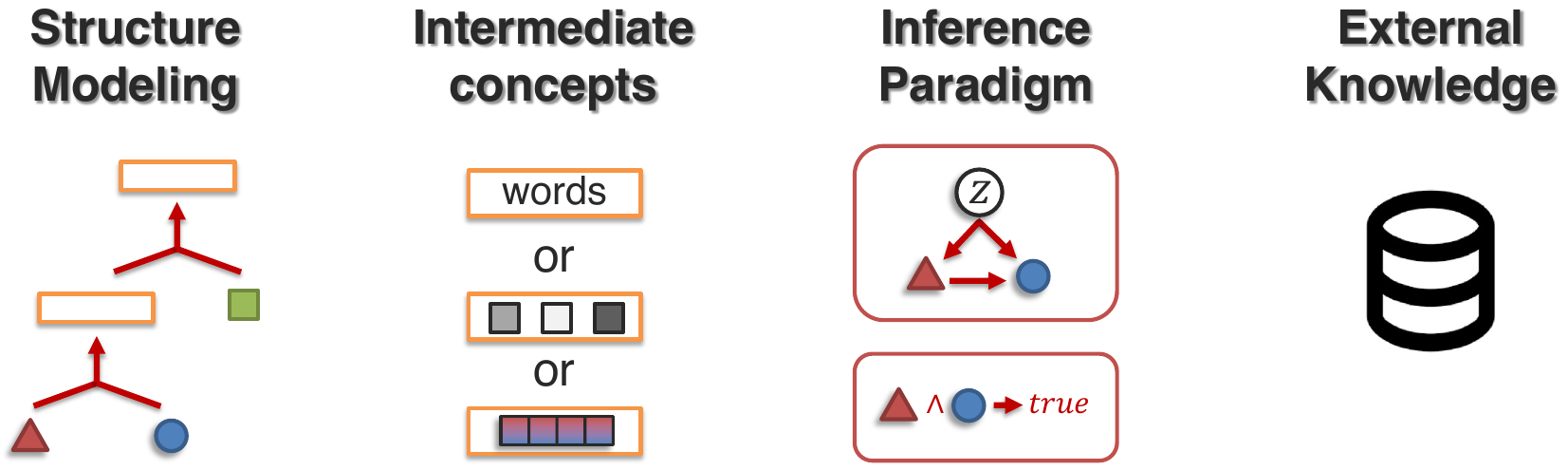}
\vspace{-2mm}
\caption{\textbf{Reasoning} aims to combine knowledge, usually through multiple inferential steps, exploiting the problem structure. Reasoning involves (1) \textit{structure modeling}: defining or learning the relationships over which reasoning occurs, (2) the \textit{intermediate concepts} used in reasoning, (3) \textit{inference} of increasingly abstract concepts from evidence, and (4) leveraging \textit{external knowledge} in the study of structure, concepts, and inference.}
\label{fig:reason}
\vspace{-4mm}
\end{figure}

\vspace{-2mm}
\section{Challenge 3: Reasoning}
\label{sec:reasoning}
\vspace{-1mm}

Reasoning is defined as combining knowledge, usually through multiple inferential steps, exploiting multimodal alignment and the problem structure. We categorize work towards multimodal reasoning into $4$ subchallenges of structure modeling, intermediate concepts, inference paradigm, and external knowledge (Figure~\cref{fig:reason}). (1) \textit{Structure modeling} involves defining or learning the relationships over which reasoning occurs, (2) \textit{intermediate concepts} studies the parameterization of individual multimodal concepts in the reasoning process, (3) \textit{inference paradigm} learns how increasingly abstract concepts are inferred from individual multimodal evidence, and (4) \textit{external knowledge} aims to leverage large-scale databases in the study of structure, concepts, and inference.

\vspace{-2mm}
\subsection{Subchallenge 3a: Structure Modeling}
\label{sec:reasoning1}
\vspace{-1mm}

\begin{figure}[t]
\centering
\vspace{-0mm}
\includegraphics[width=0.6\linewidth]{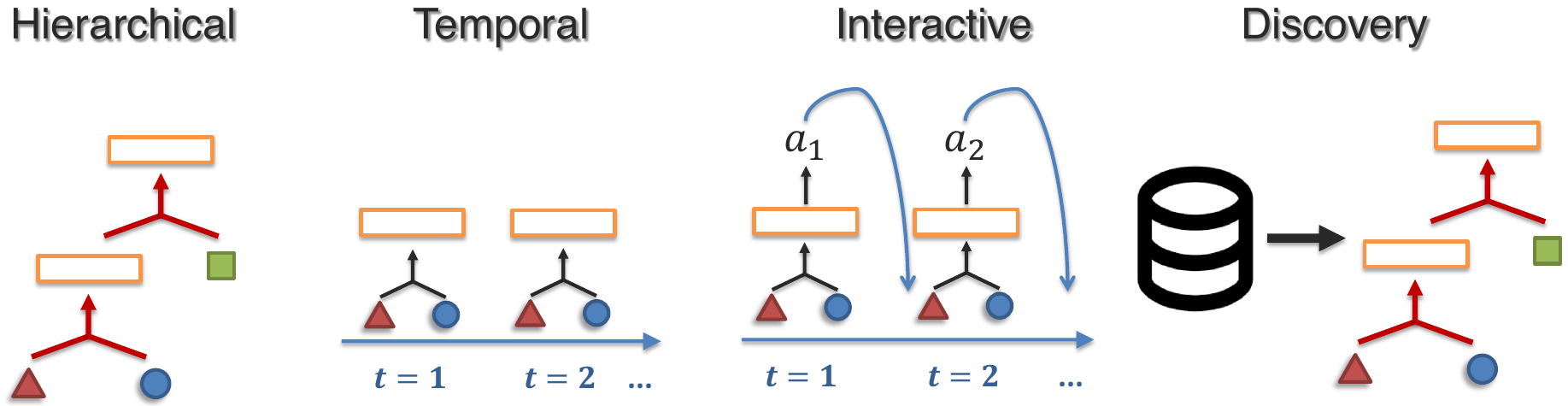}
\vspace{-2mm}
\caption{\textbf{Structure modeling} aims to define the relationship over which composition occurs, which can be (1) \textit{hierarchical} (i.e., more abstract concepts are defined as a function of less abstract ones), (2) \textit{temporal} (i.e., organized across time), (3) \textit{interactive} (i.e., where the state changes depending on each step's decision), and (4) \textit{discovered} when the latent structure is unknown and instead directly inferred from data and optimization.}
\label{fig:reason1}
\vspace{-4mm}
\end{figure}

Structure modeling aims to capture the hierarchical relationship over which composition occurs, usually via a data structure parameterizing atoms, relations, and the reasoning process. Commonly used data structures include trees~\cite{hong2019learning}, graphs~\cite{Yu2019HeterogeneousGL}, or neural modules~\cite{andreas2016neural}. We cover recent work in modeling latent \textit{hierarchical}, \textit{temporal}, and \textit{interactive} structure, as well as \textit{structure discovery} when the latent structure is unknown (Figure~\cref{fig:reason1}).

\textbf{Hierarchical structure} defines a system of organization where abstract concepts are defined as a function of less abstract ones. Hierarchical structure is present in many tasks involving language syntax, visual syntax, or higher-order reasoning. These approaches typically construct a graph based on predefined node and edge categories before using (heterogeneous variants of) graph neural networks to capture a representation of structure~\cite{shi2016survey}, such as using language syntactic structure to guide visual modules that discover specific information in images~\cite{andreas2016neural,cirik2018using}. Graph-based reasoning approaches have been applied for visual commonsense reasoning~\cite{lin2019kagnet}, visual question answering~\cite{saqur2020multimodal}, machine translation~\cite{yin2020novel}, recommendation systems~\cite{tao2020mgat}, web image search~\cite{wang2012multimodal}, and social media analysis~\cite{schinas2015multimodal}.

\textbf{Temporal structure} extends the notion of compositionality to elements across time, which is necessary when modalities contain temporal information, such as in video, audio, or time-series data. Explicit memory mechanisms have emerged as a popular choice to accumulate multimodal information across time so that long-range cross-modal interactions can be captured through storage and retrieval from memory.~\citet{rajagopalan2016extending} explore various memory representations including multimodal fusion, coordination, and factorization. Insights from key-value memory~\cite{xiong2016dynamic} and attention-based memory~\cite{zadeh2018memory} have also been successfully applied to applications including question answering, video captioning, emotion recognition, and sentiment analysis.

\textbf{Interactive structure} extends the challenge of reasoning to interactive settings, where the state of the reasoning agent changes depending on the local decisions made at every step. Typically formalized by the sequential decision-making framework, the challenge lies in maximizing long-term cumulative reward despite only interacting with the environment through short-term actions~\cite{sutton2018reinforcement}. To tackle the challenges of interactive reasoning, the growing research field of multimodal reinforcement learning (RL) has emerged from the intersection of language understanding, embodiment in the visual world, deep reinforcement learning, and robotics. We refer the reader to the extensive survey paper by~\citet{luketina2019survey} and the position paper by~\citet{bisk2020experience} for a full review of this field.~\citet{luketina2019survey} separate the literature into multimodal-conditional RL (in which multimodal interaction is necessitated by the problem formulation itself, such as instruction following~\cite{chaplot2017gated,wang2019reinforced}) and language-assisted RL (in which multimodal data is optionally used to facilitate learning, such as reading instruction manuals~\cite{narasimhan2018grounding}).

\textbf{Structure discovery}: It may be challenging to define the structure of multimodal composition without some domain knowledge of the given task. As an alternative approach, recent work has also explored using differentiable strategies to automatically search for the structure in a fully data-driven manner. To do so, one first needs to define a candidate set of reasoning atoms and relationships, before using a `meta' approach such as architecture search to automatically search for the ideal sequence of compositions for a given task~\cite{perez2019mfas,xu2021mufasa}. These approaches can benefit from optimization tricks often used in the neural architecture search literature. Memory, Attention, and Composition (MAC) similarly search for a series of attention-based reasoning steps from data in an end-to-end approach~\cite{hudson2018compositional}. Finally,~\citet{hu2017learning} extend the predefined reasoning structure obtained through language parsing in~\citet{andreas2016neural} by instead using policy gradients to automatically optimize a compositional structure over a discrete set of neural modules.

\vspace{-2mm}
\subsection{Subchallenge 3b: Intermediate Concepts}
\label{sec:reasoning2}
\vspace{-1mm}

The second subchallenge studies how we can parameterize individual multimodal concepts within the reasoning process. While intermediate concepts are usually dense vector representations in standard neural architectures, there has also been substantial work towards interpretable attention maps, discrete symbols, and language as an intermediate medium for reasoning.

\textbf{Attention maps} are a popular choice for intermediate concepts since they are, to a certain extent, human-interpretable, while retaining differentiability. For example,~\citet{andreas2016neural} design individual modules such as `attend', `combine', `count', and `measure' that are each parametrized by attention operations on the input image for visual question answering.~\citet{xu2015show} explore both soft and hard attention mechanisms for reasoning in image captioning generation. Related work has also used attention maps through dual attention architectures~\citep{nam2017dual} or stacked latent attention architectures~\citep{fan2018stacked} for multimodal reasoning. These are typically applied for problems involving complex reasoning steps such as CLEVR~\citep{johnson2017clevr} or VQA~\citep{zhang2020multimodal}.

\textbf{Discrete symbols}: A further level of discretization beyond attention maps involves using discrete symbols to represent intermediate concepts. Recent work in neuro-symbolic learning aims to integrate these discrete symbols as intermediate steps in multimodal reasoning in tasks such as visual question answering~\cite{andreas2016neural,mao2018neuro,vedantam2019probabilistic} or referring expression recognition~\cite{cirik2018using}. A core challenge in this approach lies in maintaining differentiability of discrete symbols, which has been tackled via logic-based differentiable reasoning~\cite{amizadeh2020neuro,serafini2016logic}.

\textbf{Language as a medium}: Finally, perhaps the most human-understandable form of intermediate concepts is natural language (through discrete words or phrases) as a medium. Recently,~\citet{zeng2022socratic} explore using language as an intermediate medium to coordinate multiple separate pretrained models in a zero-shot manner. Several approaches also used language phrases obtained from external knowledge graphs to facilitate interpretable reasoning~\citep{gui2021kat,zhu2015building}.~\citet{hudson2019learning} designed a neural state machine to simulate the execution of a question being asked about an image, while using discrete words as intermediate concepts.

\vspace{-2mm}
\subsection{Subchallenge 3c: Inference Paradigms}
\label{sec:reasoning3}
\vspace{-1mm}

The third subchallenge in multimodal reasoning defines the way in which increasingly abstract concepts are inferred from individual multimodal evidence. While advances in local representation fusion (such as additive, multiplicative, tensor-based, attention-based, and sequential fusion, see \S\cref{sec:representation1} for a full review) are also generally applicable here, the goal is reasoning is to be more interpretable in the inference process through domain knowledge about the multimodal problem. To that end, we cover recent directions in explicitly modeling the inference process via logical and causal operators as examples of recent trends in this direction.

\textbf{Logical inference}: Logic-based differentiable reasoning has been widely used to represent knowledge in neural networks~\cite{amizadeh2020neuro,serafini2016logic}. Many of these approaches use differentiable fuzzy logic~\cite{van2022analyzing} which provides a probabilistic interpretation of logical predicates, functions, and constants to ensure differentiability. These logical operators have been applied for visual question answering~\cite{gokhale2020vqa} and visual reasoning~\cite{amizadeh2020neuro}. Among the greatest benefits of logical reasoning lies in its ability to perform interpretable and compositional multi-step reasoning~\cite{hudson2019gqa}. Logical frameworks have also been useful for visual-textual entailment~\cite{suzuki2019multimodal} and geometric numerical reasoning~\cite{chen2021geoqa}, fields where logical inductive biases are crucial toward strong performance.

\textbf{Causal inference} extends the associational level of reasoning to interventional and counterfactual levels~\cite{pearl2009causality}, which requires extensive knowledge of the world to imagine counterfactual effects. For example,~\citet{yi2019clevrer} propose the CLEVRER benchmark focusing on four specific elements of reasoning on videos: descriptive (e.g., ‘what color’), explanatory (‘what’s responsible for’), predictive (‘what will happen next’), and counterfactual (‘what if’). Beyond CLEVRER, recent work has also proposed Causal VQA~\cite{agarwal2020towards} and Counterfactual VQA~\cite{niu2021counterfactual} to measure the robustness of VQA models under controlled interventions to the question as a step towards mitigating language bias in VQA models. Methods inspired by integrating causal reasoning capabilities into neural network models have also been shown to improve robustness and reduce biases~\cite{wang2020visual}.

\vspace{-2mm}
\subsection{Subchallenge 3d: External Knowledge}
\label{sec:reasoning4}
\vspace{-1mm}

The final subchallenge studies the derivation of knowledge in the study of defining composition and structure. Knowledge is typically derived from domain knowledge on task-specific datasets. As an alternative to using domain knowledge to pre-define the compositional structure, recent work has also explored reasoning automatically using data-driven methods, such as widely accessible but more weakly supervised data outside the immediate task domain.

\textbf{Multimodal knowledge graphs} extend classic work in language and symbolic knowledge graphs (e.g., Freebase~\cite{bollacker2008freebase}, DBpedia~\cite{auer2007dbpedia}, YAGO~\cite{suchanek2007yago}, WordNet~\cite{miller1995wordnet}) to semantic networks containing multimodal concepts as nodes and multimodal relationships as edges~\cite{zhu2022multi}. Multimodal knowledge graphs are important because they enable the grounding of structured information in the visual and physical world. For example,~\citet{liu2019mmkg} constructs multimodal knowledge graphs containing both numerical features and images for entities. Visual Genome is another example containing dense annotations of objects, attributes, and relationships in images and text~\cite{krishna2017visual}. These multimodal knowledge bases have been shown to benefit visual question answering~\cite{wu2016ask,zhu2015building}, knowledge base completion~\cite{pezeshkpour2018embedding}, and image captioning~\cite{melas2018training,yao2018exploring}.~\citet{gui2021kat} integrates knowledge into vision-and-language transformers for automatic reasoning over both knowledge sources. We refer the reader to a comprehensive survey by~\citet{zhu2022multi} for additional references.

\textbf{Multimodal commonsense reasoning} requires deeper real-world knowledge potentially spanning logical, causal, and temporal relationships between concepts. For example, elements of causal reasoning are required to answer the questions regarding images in VCR~\cite{zellers2019vcr} and VisualCOMET~\cite{park2020visualcomet}, while other works have also introduced datasets with video and text inputs to test for temporal reasoning (e.g., MovieQA~\cite{tapaswi2016movieqa}, MovieFIB~\cite{maharaj2017dataset}, TVQA~\cite{lei2018tvqa}). Benchmarks for multimodal commonsense typically require leveraging external knowledge from knowledge bases~\cite{song2021kvl} or pretraining paradigms on large-scale datasets~\cite{lu2019vilbert,zellers2021merlot}.
\vspace{-2mm}
\section{Challenge 4: Generation}
\label{sec:generation}
\vspace{-1mm}

\begin{figure}[t]
\centering
\vspace{-0mm}
\includegraphics[width=0.6\linewidth]{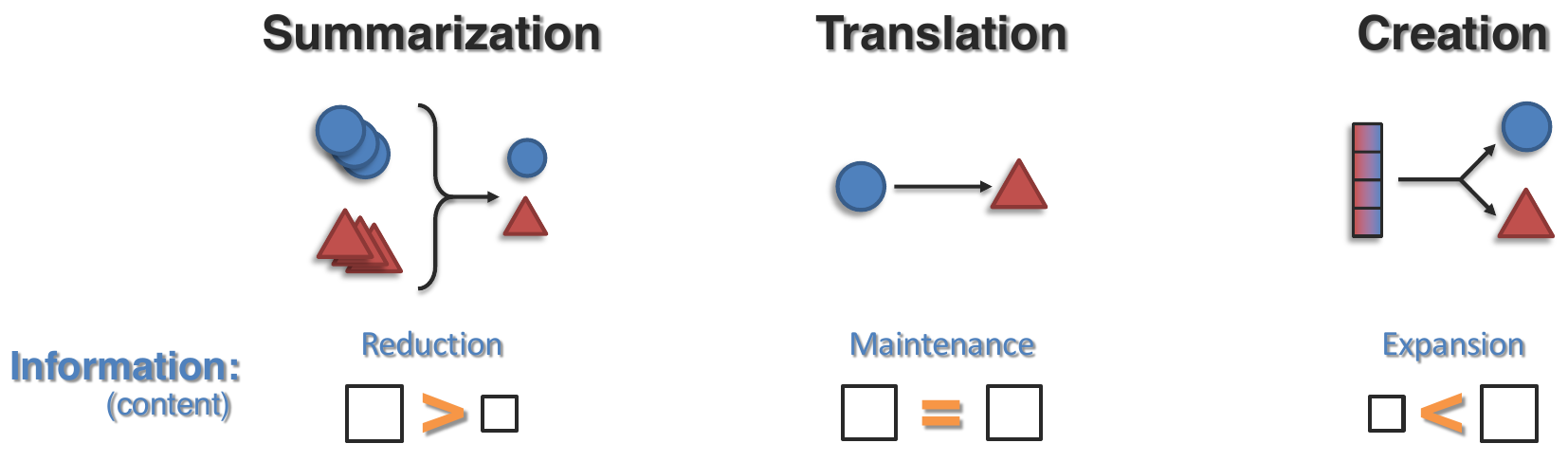}
\vspace{-2mm}
\caption{How can we learn a generative process to produce raw modalities that reflect cross-modal interactions, structure, and coherence? \textbf{Generation} involves (1) \textit{summarizing} multimodal data to highlight the most salient parts, (2) \textit{translating} from one modality to another while being consistent with modality connections, and (3) \textit{creating} multiple modalities simultaneously while maintaining coherence.}
\label{fig:gen}
\vspace{-4mm}
\end{figure}

The fourth challenge involves learning a generative process to produce raw modalities that reflect cross-modal interactions, structure, and coherence, through \textit{summarization}, \textit{translation}, and \textit{creation} (Figure~\cref{fig:gen}). These three categories are distinguished based on the information change from input to output modalities, following categorizations in text generation~\citep{deng2021compression}. We will cover recent advances as well as the evaluation of generated content.

\vspace{-2mm}
\subsection{Subchallenge 4a: Summarization}
\label{sec:generation1}
\vspace{-1mm}

Summarization aims to compress data to create an abstract that represents the most important or relevant information within the original content. Recent work has explored various input modalities to guide text summarization, such as images~\cite{chen2018abstractive}, video~\cite{li2020vmsmo}, and audio~\cite{evangelopoulos2013multimodal,jangra2020text,li2017multi}. Recent trends in multimodal summarization include \textit{extractive} and \textit{abstractive} approaches. Extractive approaches aim to filter words, phrases, and other unimodal elements from the input to create a summary~\cite{chen2018extractive,jangra2020text,li2017multi}. Beyond text as output, video summarization is the task of producing a compact version of the video (visual summary) by encapsulating the most informative parts~\cite{sah2017semantic}.~\citet{li2017multi} collected a dataset of news videos and articles paired with manually annotated summaries as a benchmark towards multimodal summarization. Finally,~\citet{uzzaman2011multimodal} aim to simplify complex sentences by extracting multimodal summaries for accessibility. On the other hand, abstractive approaches define a generative model to generate the summary at multiple levels of granularity~\cite{chen2018abstractive,li2019keep}. Although most approaches only focus on generating a textual summary from multimodal data~\cite{palaskar2019multimodal}, several directions have also explored generating summarized images to supplement the generated textual summary~\cite{chen2018abstractive,li2020vmsmo}.

\vspace{-2mm}
\subsection{Subchallenge 4b: Translation}
\label{sec:generation2}
\vspace{-1mm}

Translation aims to map one modality to another while respecting semantic connections and information content~\citep{vinyals2016show}. For example, generating a descriptive caption of an image can help improve the accessibility of visual content for blind people~\citep{gurari2018vizwiz}. Multimodal translation brings about new difficulties involving the generation of high-dimensional structured data as well as their evaluation. Recent approaches can be classified as \textit{exemplar-based}, which are limited to retrieving from training instances to translate between modalities but guarantee fidelity~\cite{farhadi2010every}, and \textit{generative} models which can translate into arbitrary instances interpolating beyond the data but face challenges in quality, diversity, and evaluation~\cite{koh2021text,ramesh2021zero,tsimpoukelli2021multimodal}. Despite these challenges, recent progress in large-scale translation models has yielded impressive quality of generated content in text-to-image~\citep{rombach2022high,ramesh2021zero}, text-to-video~\citep{singer2022make}, audio-to-image~\cite{jamaludin2019you}, text-to-speech~\cite{ren2019fastspeech}, speech-to-gesture~\citep{ahuja2020style}, speaker-to-listener~\citep{ng2022learning}, language to pose~\citep{ahuja2019language2pose}, and speech and music generation~\cite{oord2018parallel}.

\vspace{-2mm}
\subsection{Subchallenge 4c: Creation}
\label{sec:generation3}
\vspace{-1mm}

Creation aims to generate novel high-dimensional data (which could span text, images, audio, video, and other modalities) from small initial examples or latent conditional variables. This \textit{conditional decoding} process is extremely challenging since it needs to be (1) conditional: preserve semantically meaningful mappings from the initial seed to a series of long-range parallel modalities, (2) synchronized: semantically coherent across modalities, (3) stochastic: capture many possible future generations given a particular state, and (4) auto-regressive across possibly long ranges.
Many modalities have been considered as targets for creation.
Language generation has been explored for a long time~\cite{radford2019language}, and recent work has explored high-resolution speech and sound generation using neural networks~\cite{oord2018parallel}. Photorealistic image generation has also recently become possible due to advances in large-scale generative modeling~\cite{karras2020analyzing}. Furthermore, there have been a number of attempts at generating abstract scenes~\cite{tan2019text2scene}, computer graphics~\cite{mildenhall2020nerf}, and talking heads~\cite{zhu2021arbitrary}. While there has been some progress toward video generation~\citep{singer2022make}, complete synchronized generation of realistic video, text, and audio remains a challenge.

Finally, one of the biggest challenges facing multimodal generation is difficulty in evaluating generated content, especially when there exist serious ethical issues when fake news~\citep{bender2021dangers}, hate speech~\cite{gehman2020realtoxicityprompts,abid2021persistent}, deepfakes~\cite{hancock2021social}, and lip-syncing videos~\citep{suwajanakorn2017synthesizing} can be easily generated. 
While the ideal way to evaluate generated content is through user studies, it is time-consuming, costly, and can potentially introduce subjectivity bias into the evaluation process~\citep{geva2019we}. Several automatic proxy metrics have been proposed~\cite{anderson2016spice,chen2020comprises} by none are universally robust across many generation tasks.
\vspace{-2mm}
\section{Challenge 5: Transference}
\label{sec:transference}
\vspace{-1mm}

\begin{figure}[t]
\centering
\vspace{-0mm}
\includegraphics[width=0.6\linewidth]{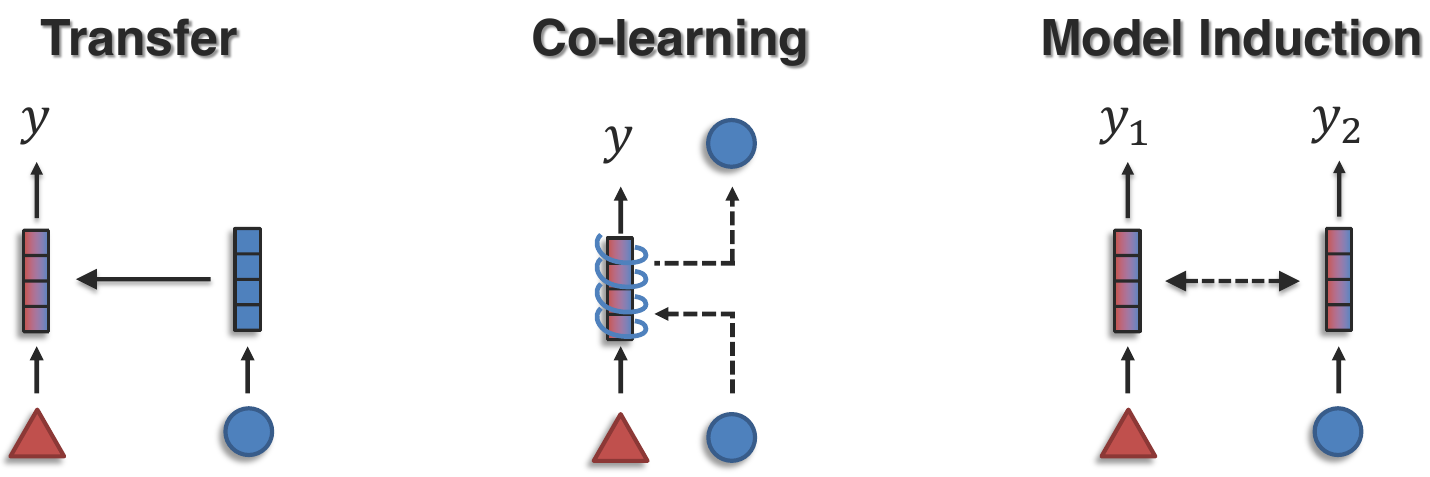}
\vspace{-2mm}
\caption{\textbf{Transference} studies the transfer of knowledge between modalities, usually to help a noisy or limited primary modality, via (1) \textit{cross-modal transfer} from models trained with abundant data in the secondary modality, (2) \textit{multimodal co-learning} to share information across modalities by sharing representations, and (3) \textit{model induction} that keeps individual unimodal models separate but induces behavior in separate models.}
\label{fig:transference}
\vspace{-4mm}
\end{figure}

Transference aims to transfer knowledge between modalities and their representations. How can knowledge learned from a secondary modality (e.g., predicted labels or representation) help a model trained on a primary modality? This challenge is particularly relevant when the primary modality has limited resources — a lack of annotated data, noisy inputs, or unreliable labels. We call this challenge transference since the transfer of information from the secondary modality gives rise to new behaviors previously unseen in the primary modality. We identify three types of transference approaches: (1) \textit{cross-modal transfer}, (2) \textit{multimodal co-learning}, and (3) \textit{model induction} (Figure~\cref{fig:transference}).

\vspace{-2mm}
\subsection{Subchallenge 5a: Cross-modal Transfer}
\label{sec:transference1}
\vspace{-1mm}

In most settings, it may be easier to collect either labeled or unlabeled data in the secondary modality and train strong supervised or pretrained models. These models can then be conditioned or fine-tuned for a downstream task involving the primary modality. In other words, this line of research extends unimodal transfer and fine-tuning to cross-modal settings.

\textbf{Tuning}: Inspired by prior work in NLP involving prefix tuning~\cite{li2021prefix} and prompt tuning~\cite{lester2021power}, recent work has also studied the tuning of pretrained language models to condition on visual and other modalities. For example,~\citet{tsimpoukelli2021multimodal} quickly conditions a pretrained, frozen language model on images for image captioning. Related work has also adapted prefix tuning for image captioning~\cite{chen2021visualgpt}, multimodal fusion~\cite{hasan2021humor}, and summarization~\cite{yu2021vision}. While prefix tuning is simple and efficient, it provides the user with only limited control over how information is transferred. Representation tuning goes a level deeper by modifying the inner representations of the language model via contextualization with other modalities. For example,~\citet{ziegler2019encoder} includes additional self-attention layers between language model layers and external modalities.~\citet{rahman2020integrating} design a shifting gate to adapt language model layers with audio and visual information.

\textbf{Multitask learning} aims to use multiple large-scale tasks to improve performance as compared to learning on individual tasks. Several models such as Perceiver~\cite{jaegle2021perceiver}, MultiModel~\cite{kaiser2017one}, ViT-BERT~\cite{li2021towards}, and PolyViT~\cite{likhosherstov2022polyvit} have explored the possibility of using the same unimodal encoder architecture for different inputs across unimodal tasks (i.e., language, image, video, or audio-only). The Transformer architecture has emerged as a popular choice due to its suitability for serialized inputs such as text (sequence of tokens)~\cite{devlin2019bert}, images (sequence of patches)~\cite{dosovitskiy2020image}, video (sequence of images)~\cite{sun2019videobert}, and other time-series data (sequence of timesteps)~\cite{lim2021temporal}.
There have also been several attempts to build a single model that works well on a suite of multimodal tasks, including both not limited to HighMMT~\cite{liang2022highmmt}, VATT~\cite{akbari2021vatt}, FLAVA~\cite{singh2021flava}, and Gato~\citep{reed2022generalist}.

\textbf{Transfer learning}: While more research has focused on transfer within the same modality with external information~\cite{socher2013zero,NIPS2019_8731,zadeh2020foundations},~\citet{liang2021cross} studies transfer to new modalities using small amounts of paired but unlabeled data.~\citet{lu2021pretrained} found that Transformers pretrained on language transfer to other sequential modalities as well.~\citet{liang2022highmmt} builds a single multimodal model capable of transferring to completely new modalities and tasks. Recently, there has also been a line of work investigating the transfer of pretrained language models for planning~\cite{huang2022language} and interactive decision-making~\cite{li2022pre}.

\vspace{-2mm}
\subsection{Subchallenge 5b: Multimodal Co-learning}
\label{sec:transference2}
\vspace{-1mm}

Multimodal co-learning aims to transfer information learned through secondary modalities to target tasks involving the primary modality by sharing intermediate representation spaces between both modalities. These approaches essentially result in a single joint model across all modalities.

\textbf{Co-learning via representation} aims to learn either a joint or coordinated representation space using both modalities as input. Typically, this involves adding secondary modalities during the training process, designing a suitable representation space, and investigating how the multimodal model transfers to the primary modality during testing. For example, DeViSE learns a coordinated similarity space between image and text to improve image classification~\cite{frome2013devise}.~\citet{marino2016more} use knowledge graphs for image classification via a graph-based joint representation space.~\citet{jia2021scaling} improve image classifiers with contrastive representation learning between images and noisy captions.
Finally,~\citet{zadeh2020foundations} showed that implicit co-learning is also possible without explicit co-learning objectives.

\textbf{Co-learning via generation} instead learns a translation model from  the primary to secondary modality, resulting in enriched representations of the primary modality that can predict both the label and `hallucinate' secondary modalities containing shared information. Classic examples in this category includes language modeling by mapping contextualized text embeddings into images~\citep{tan2020vokenization}, image classification by projecting image embeddings into word embeddings~\citep{socher2013zero}, and language sentiment analysis by translating language into video and audio~\citep{pham2019found}.

\vspace{-2mm}
\subsection{Subchallenge 5c: Model Induction}
\label{sec:transference3}
\vspace{-1mm}

In contrast to co-learning, model induction approaches keep individual unimodal models across primary and secondary modalities separate but aim to induce behavior in both models. Model induction is exemplified by co-training, in which two learning algorithms are trained separately on each view of the data before using each algorithm's predictions to pseudo-label new unlabeled examples to enlarge the training set of the other view~\cite{blum1998combining}. Therefore, information is transferred across multiple views through model predictions instead of shared representation spaces.

\textbf{Multimodal co-training} extends co-training by jointly learning classifiers for multiple modalities~\cite{hinami2018multimodal}.~\citet{guillaumin2010multimodal} study semi-supervised learning by using a classifier on both image and text to pseudo-label unlabeled images before training a final classifier on both labeled and unlabeled images.~\citet{cheng2016semi} performs semi-supervised multimodal learning using a diversity-preserving co-training algorithm. Finally,~\citet{dunnmon2020cross} applies ideas from data programming to the problem of cross-modal weak supervision, where weak labels derived from a secondary modality (e.g., text) are used to train models over the primary modality (e.g., images).

\textbf{Co-regularization}: Another set of models employs a regularizer that penalizes functions from either modality that disagree with each other. This class of models, called co-regularization, is a useful technique to control model complexity by preferring hypothesis classes containing models that predict similarly across the two views~\citep{sindhwani2005co}.~\citet{sridharan2008information} provide guarantees for these approaches using an information-theoretic framework. More recently, similar co-regularization approaches have also been applied for multimodal feature selection~\citep{hsieh2019adaptive}, semi-supervised multimodal learning~\citep{yang2019comprehensive}, and video summarization~\citep{morere2015co}.
\begin{figure}[t]
\centering
\vspace{-0mm}
\includegraphics[width=0.7\linewidth]{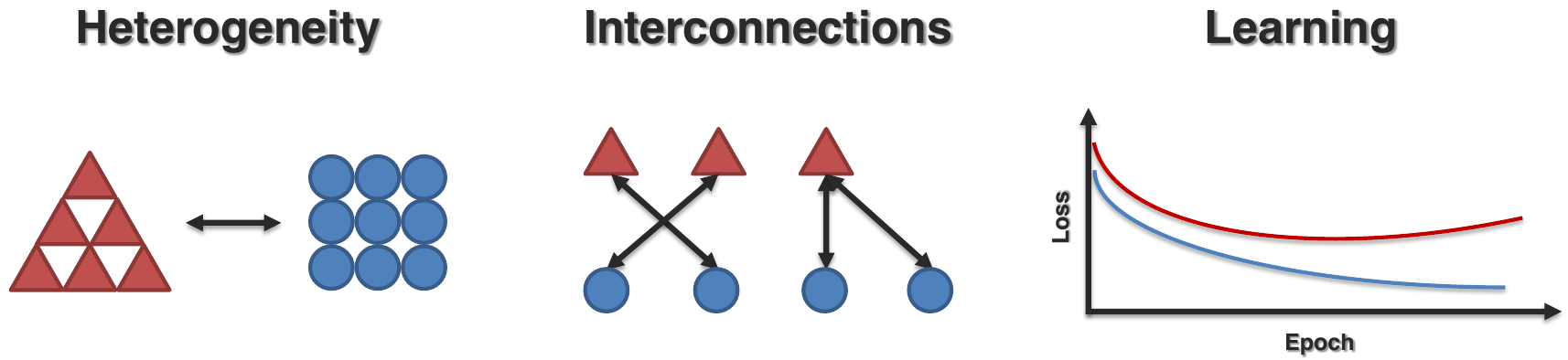}
\vspace{-2mm}
\caption{\textbf{Quantification}: what are the empirical and theoretical studies we can design to better understand (1) the dimensions of \textit{heterogeneity}, (2) the presence and type of \textit{interconnections}, and (3) the \textit{learning} and optimization challenges?}
\label{fig:quant}
\vspace{-4mm}
\end{figure}

\vspace{-2mm}
\section{Challenge 6: Quantification}
\label{sec:quantification}
\vspace{-1mm}

Quantification aims to provide a deeper empirical and theoretical study of multimodal models to gain insights and improve their robustness, interpretability, and reliability in real-world applications. We break down quantification into 3 sub-challenges: (1) quantifying the \textit{dimensions of heterogeneity} and how they subsequently influence modeling and learning, (2) quantifying the presence and type of \textit{connections and interactions} in multimodal datasets and trained models, and (3) characterizing the \textit{learning and optimization} challenges involved when learning from heterogeneous data (Figure~\cref{fig:quant}).

\vspace{-2mm}
\subsection{Subchallenge 6a: Dimensions of Heterogeneity}
\label{sec:quantification1}
\vspace{-1mm}

\begin{figure}[t]
\centering
\vspace{-0mm}
\includegraphics[width=0.5\linewidth]{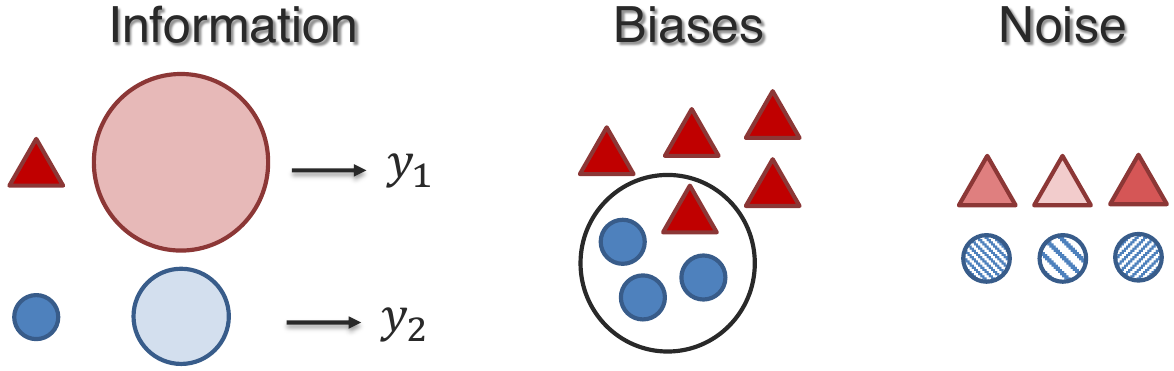}
\vspace{-2mm}
\caption{The subchallenge of \textbf{heterogeneity} quantification aims to understand the dimensions of heterogeneity commonly encountered in multimodal research, such as (1) different quantities and usages of \textit{modality information}, (2) the presence of \textit{modality biases}, and (3) quantifying and mitigating \textit{modality noise}.}
\label{fig:quant1}
\vspace{-4mm}
\end{figure}

This subchallenge aims to understand the dimensions of heterogeneity commonly encountered in multimodal research, and how they subsequently influence modeling and learning (Figure~\cref{fig:quant1}).

\textbf{Modality information}: Understanding the information of entire modalities and their constituents is important for determining which segment of each modality contributed to subsequent modeling. Recent work can be categorized into: (1) interpretable methods that explicitly model how each modality is used~\cite{park2018multimodal,tsai2020multimodal,zadeh2018multimodal} or (2) post-hoc explanations of black-box models~\cite{chandrasekaran2018explanations,goyal2016towards}. In the former, methods such as Concept Bottleneck Models~\cite{koh2020concept} and fitting sparse linear layers~\cite{wong2021leveraging} or decision trees~\cite{wan2020nbdt} on top of deep feature representations have emerged as promising choices.
In the latter, approaches such as gradient-based visualizations~\cite{simonyan2013deep,goyal2016towards,selvaraju2017grad}) and feature attributions (e.g., modality contribution~\citep{gat2021perceptual}, LIME~\cite{ribeiro2016should}, and Shapley values~\cite{merrick2020explanation}) have been used to highlight regions of each modality used by the model.

\textbf{Modality biases} are unintended correlations between input and outputs that could be introduced during data collection~\cite{birhane2021multimodal,bolukbasi2016man}, modeling~\cite{geirhos2020shortcut}, or during human annotation~\cite{devillers2005challenges}. Modality biases can lead to unexpectedly poor performance in the real world~\cite{sakaguchi2020winogrande}, or even more dangerously, potential for harm towards underrepresented groups~\citep{pena2020faircvtest,hendricks2018women}. For example,~\citet{goyal2017making} found \textit{unimodal biases} in the language modality of VQA tasks, resulting in mistakes due to ignoring visual information~\cite{agrawal2016analyzing}. Subsequent work has developed carefully-curated diagnostic benchmarks to mitigate data collection biases, like VQA 2.0~\cite{goyal2017making}, GQA~\cite{hudson2019gqa}, and NLVR2~\cite{suhr2019nlvr2}. Recent work has also found compounding \textit{social biases} in multimodal systems~\cite{ross2020measuring,srinivasan2021worst,cho2022dall} stemming from gender bias in both language and visual modalities~\citep{buolamwini2018gender,sheng2019woman}, which may cause danger when deployed~\citep{pena2020faircvtest}.

\textbf{Modality noise topologies and robustness}: The study of modality noise topologies aims to benchmark and improve how multimodal models perform in the presence of real-world data imperfections. Each modality has a unique noise topology, which determines the distribution of noise and imperfections that it commonly encounters. For example, images are susceptible to blurs and shifts, typed text is susceptible to typos following keyboard positions, and multimodal time-series data is susceptible to correlated imperfections across synchronized time steps.~\citet{liang2021multibench} collect a comprehensive set of targeted noisy distributions unique to each modality.
In addition to natural noise topologies, related work has also explored adversarial attacks~\cite{ding2021multimodal} and distribution shifts~\cite{foltyn2021towards} in multimodal systems. There has also been some progress in accounting for noisy or missing modalities by modality imputation using probabilistic models~\cite{ma2021smil}, autoencoders~\citep{DBLP:conf/cvpr/Tran0ZJ17}, translation models~\cite{pham2019found}, or low-rank approximations~\cite{liang2019tensor}. However, they run the risk of possible error compounding and require knowing which modalities are imperfect beforehand.

\vspace{-2mm}
\subsection{Subchallenge 6b: Modality Interconnections}
\label{sec:quantification2}
\vspace{-1mm}

\begin{figure}[t]
\centering
\vspace{-0mm}
\includegraphics[width=0.5\linewidth]{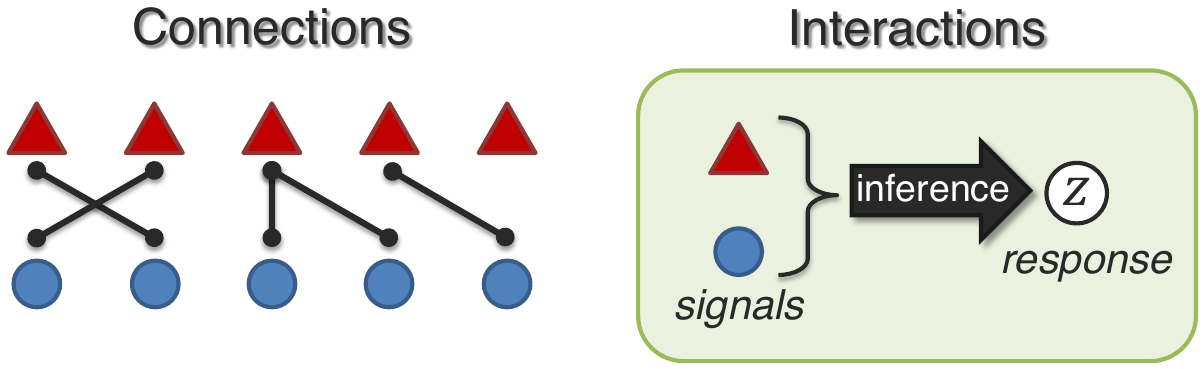}
\vspace{-2mm}
\caption{Quantifying \textbf{modality interconnections} studies (1) \textit{connections}: can we discover what modality elements are related to each other and why, and (2) \textit{interactions}: can we understand how modality elements interact during inference?}
\label{fig:quant2}
\vspace{-4mm}
\end{figure}

Modality connections and interactions are an essential component of multimodal models, which has inspired an important line of work in visualizing and understanding the nature of modality interconnections in datasets and trained models. We divide recent work into quantification of (1) \textit{connections}: how modalities are related and share commonality, and (2) \textit{interactions}: how modality elements interact during inference (Figure~\cref{fig:quant2}).

\textbf{Connections}: Recent work has explored the quantification of modality connections through visualization tools on joint representation spaces~\citep{Itkina2020EvidentialSO} or attention maps~\citep{aflalo2022vl}. Perturbation-based analysis perturbs the input and observes changes in the output to understand internal connections~\citep{liang2022multiviz,niu2021counterfactual}. Finally, specifically curated diagnostic datasets are also useful in understanding semantic connections: Winoground~\citep{thrush2022winoground} probes vision and language models for visio-linguistic compositionality, and PaintSkills~\citep{cho2022dall} measures the connections necessary for visual reasoning.

\textbf{Interactions}: One common categorization of interactions involves redundancy, uniqueness, and synergy~\citep{williams2010nonnegative}. Redundancy describes task-relevant information shared among features, uniqueness studies the task-relevant information present in only one of the features, and synergy investigates the emergence of new information when both features are present. From a statistical perspective, measures of redundancy include mutual information~\cite{blum1998combining,balcan2004co} and contrastive learning estimators~\cite{tosh2021contrastive,tsai2020self}. Other approaches have studied these measures in isolation, such as redundancy via distance between prediction logits using either feature~\cite{mazzetto21a}, statistical distribution tests on input features~\cite{auffarth2010comparison}, or via human annotations~\cite{ruiz2006examining}.
From the semantic view, recent work in Causal VQA~\cite{agarwal2020towards} and Counterfactual VQA~\cite{niu2021counterfactual} seek to understand the interactions captured by trained models by measuring their robustness under controlled semantic edits to the question or image. Finally, recent work has formalized definitions of non-additive interactions to quantify their presence in trained models~\citep{sorokina2008detecting,tsang2018detecting}. Parallel research such as EMAP~\cite{hessel2020emap}, DIME~\cite{lyu2022dime}, M2Lens~\cite{wang2021m2lens}, and MultiViz~\citep{liang2022multiviz} aims to quantify the interactions in real-world multimodal datasets and models.

\vspace{-2mm}
\subsection{Subchallenge 6c: Multimodal Learning Process}
\label{sec:quantification3}
\vspace{-1mm}

\begin{figure}[t]
\centering
\vspace{-0mm}
\includegraphics[width=0.7\linewidth]{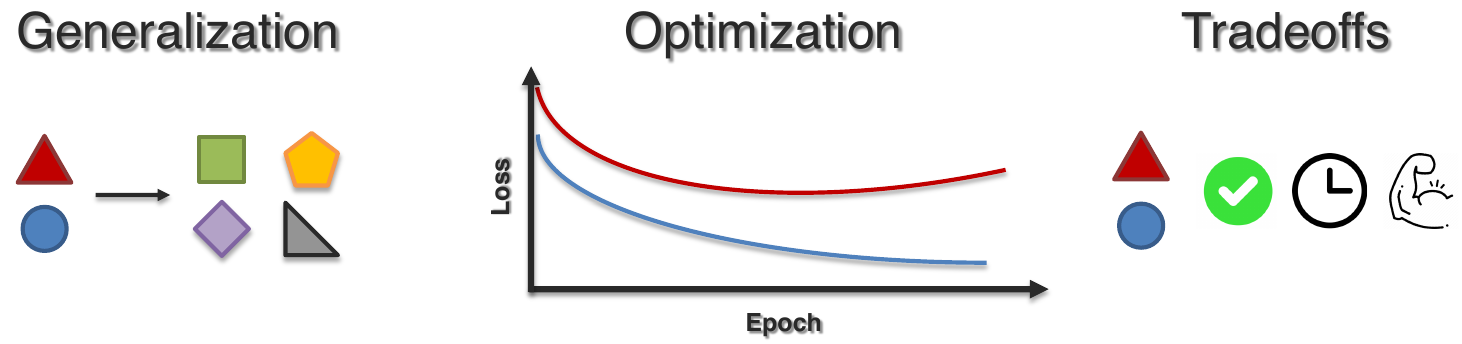}
\vspace{-2mm}
\caption{Studying the multimodal \textbf{learning process} involves understanding (1) \textit{generalization} across modalities and tasks, (2) \textit{optimization} for balanced and efficient training, and (3) \textit{tradeoffs} between performance, robustness, and complexity in the real-world deployment of multimodal models.}
\label{fig:quant3}
\vspace{-4mm}
\end{figure}

Finally, there is a need to characterize the learning and optimization challenges involved when learning from heterogeneous data. This section covers recent work in (1) \textit{generalization} across modalities and tasks, (2) better \textit{optimization} for balanced and efficient training, and (3) balancing the \textit{tradeoffs} between performance, robustness, and complexity in real-world deployment (Figure~\cref{fig:quant3}).

\textbf{Generalization}: With advances in sensing technologies, many real-world platforms such as cellphones, smart devices, self-driving cars, healthcare technologies, and robots now integrate a much larger number of sensors beyond the prototypical text, video, and audio modalities~\cite{huang2019multimodal}. Recent work has studied generalization across paired modality inputs~\citep{liang2021cross,radford2021learning} and in unpaired scenarios where each task is defined over only a small subset of all modalities~\cite{liang2022highmmt,lu2021pretrained,reed2022generalist}.

\textbf{Optimization challenges}: Related work has also explored the optimization challenges of multimodal learning, where multimodal networks are often prone to overfitting due to increased capacity, and different modalities overfit and generalize at different rates so training them jointly with a single optimization strategy is sub-optimal~\cite{wang2020makes}. Subsequent work has suggested both empirical and theoretical studies of why joint training of multimodal networks may be difficult and has proposed methods to improve the optimization process via weighting approaches~\cite{wu2022characterizing}.

\textbf{Modality Tradeoffs}: In real-world deployment, a balance between performance, robustness, and complexity is often required. Therefore, one often needs to balance the utility of additional modalities with the additional complexity in data collection and modeling~\cite{liang2021multibench} as well as increased susceptibility to noise and imperfection in the additional modality~\cite{pham2019found}.
How can we formally quantify the utility and risks of each input modality, while balancing these tradeoffs for reliable real-world usage? There have been several attempts toward formalizing the semantics of a multimodal representation and how these benefits can transfer to downstream tasks~\cite{liang2022brainish,tsai2020self,thomason2016learning}, while information-theoretic arguments have also provided useful insights~\cite{blum1998combining,sridharan2008information}.

\vspace{-2mm}
\section{Conclusion}
\vspace{-1mm}

This paper defined three core principles of modality heterogeneity, connections, and interactions central to multimodal machine learning research, before proposing a taxonomy of six core technical challenges: representation, alignment, reasoning, generation, transference, and quantification covering historical and recent directions. Despite the immense opportunities afforded by recent progress in multimodal machine learning, there remain many unsolved challenges from theoretical, computational, and application perspectives:

\vspace{-2mm}
\subsection{Future Directions}
\label{sec:future}
\vspace{-1mm}

\textbf{Representation}: \textit{Theoretical and empirical frameworks}. How can we formally define the three core principles of heterogeneity, connections, and interactions? What mathematical or empirical frameworks will enable us to taxonomize the dimensions of heterogeneity and interconnections, and subsequently quantify their presence in multimodal datasets and models? Answering these fundamental questions will lead to a better understanding of the capabilities and limitations of current multimodal representations. \textit{Beyond additive and multiplicative cross-modal interactions}. While recent work has been successful at modeling multiplicative interactions of increasing order, how can we capture causal, logical, and temporal connections and interactions? What is the right type of data and domain knowledge necessary to model these interactions? \textit{Brain and multimodal perception}. There are many core insights regarding multimodal processing to be gained from the brain and human cognition, including the brain's neural architecture~\citep{blum2021theoretical}, intrinsic multimodal properties~\citep{kosslyn2010multimodal}, mental imagery~\citep{nanay2018multimodal}, and the nature of neural signals~\citep{palazzo2020decoding}. How does the human brain represent different modalities, how is multisensory integration performed, and how can these insights inform multimodal learning? In the other direction, what are several challenges and opportunities in processing high-resolution brain signals such as fMRI and MEG/EEG, and how can multimodal learning help in the future analysis of data collected in neuroscience?

\textbf{Alignment}: \textit{Memory and long-term interactions}. Many current multimodal benchmarks only have a short temporal dimension, which has limited the demand for models that can accurately process long-range sequences and learn long-range interactions. Capturing long-term interactions presents challenges since it is difficult to semantically relate information when they occur very far apart in time or space and raises complexity issues. How can we design models (perhaps with memory mechanisms) to ensure that these long-term cross-modal interactions are captured?

\textbf{Reasoning}: \textit{Multimodal compositionality}. How can we understand the reasoning process of trained models, especially in regard to how they combine information from modality elements? This challenge of compositional generalization is difficult since many compositions of elements are typically not present during training, and the possible number of compositions increases exponentially with the number of elements~\citep{thrush2022winoground}. How can we best test for compositionality, and what reasoning approaches can enable compositional generalization?

\textbf{Transference}: \textit{High-modality learning} aims to learn representations from an especially large number of heterogeneous data sources, which is a common feature of many real-world multimodal systems such as self-driving cars and IoT~\cite{huang2019multimodal}. More modalities introduce more dimensions of heterogeneity, incur complexity challenges in unimodal and multimodal processing, and require dealing with non-parallel data (i.e., not all modalities are present at the same time).

\textbf{Generation}: \textit{Creation and real-world ethical concerns}. The complete synchronized creation of realistic video, text, and audio remains a challenge. Furthermore, the recent success in modality generation has brought ethical concerns regarding their use. For example, large-scale pretrained language models can potentially generate text denigrating to particular social groups~\cite{sheng2019woman}, toxic speech~\cite{gehman2020realtoxicityprompts}, and sensitive pretraining data~\cite{carlini2021extracting}. Future work should study how these risks are potentially amplified or reduced when the dataset is multimodal, and whether there are ethical issues specific to multimodal generation.

\textbf{Quantification}: \textit{Modality utility, tradeoffs, and selection}. How can we formalize why modalities can be useful for a task, and the potential reasons a modality can be harmful? Can we come up with formal guidelines to compare these tradeoffs and subsequently select modalities? \textit{Explainability and interpretability}. Before models can be safely used by real-world stakeholders such as doctors, educators, or policymakers, we need to understand the taxonomy of multimodal phenomena in datasets and trained models we should aim to interpret. How can we evaluate whether these phenomena are accurately interpreted? These challenges are exacerbated for relatively understudied modalities beyond language and vision, where the modalities themselves are not easy to visualize. Finally, how can we tailor these explanations, possibly in a \textit{human-in-the-loop} manner, to inform real-world decision-making? There are also core challenges in understanding and quantifying \textit{modality and social biases} as well as \textit{robustness} to imperfect, noisy, and out-of-distribution modalities.

In conclusion, we believe that our taxonomy will help to catalog future research papers and better understand the remaining unresolved problems in multimodal machine learning.

\vspace{-2mm}
\section*{Acknowledgements}
\vspace{-1mm}

This material is based upon work partially supported by the National Science Foundation (Awards \#1722822 and \#1750439), National Institutes of Health (Awards \#R01MH125740, \#R01MH096951, and \#U01MH116925), BMW of North America, and Meta.
PPL is partially supported by a Facebook PhD Fellowship and a Carnegie Mellon University's Center for Machine Learning and Health Fellowship.
Any opinions, findings, conclusions, or recommendations expressed in this material are those of the author(s) and do not necessarily reflect the views of the National Science Foundation, National Institutes of Health,  BMW of North America, Facebook, or Carnegie Mellon University's Center for Machine Learning and Health, and no official endorsement should be inferred. We are extremely grateful to Alex Wilf, Arav Agarwal, Catherine Cheng, Chaitanya Ahuja, Daniel Fried, Dong Won Lee, Jack Hessel, Leena Mathur, Lenore Blum, Manuel Blum, Martin Ma, Peter Wu, Richard Chen, Ruslan Salakhutdinov, Santiago Benoit, Su Min Park, Torsten Wortwein, Victoria Lin, Volkan Cirik, Yao-Hung Hubert Tsai, Yejin Choi, Yiwei Lyu, Yonatan Bisk, and Youssouf Kebe for helpful discussions and feedback on initial versions of this paper.

{\footnotesize
\bibliographystyle{ACM-Reference-Format}
\bibliography{refs}
}

\end{document}